\documentclass[letterpaper]{article} \usepackage{aaai25}
\usepackage{times}  \usepackage{helvet}  \usepackage{courier}  \usepackage[hyphens]{url}  \usepackage{graphicx}    \usepackage{natbib}  \usepackage{caption}    

\usepackage{algorithm}
\usepackage{algpseudocode}
\usepackage{newfloat}
\usepackage{listings}
\usepackage{booktabs}
\usepackage[table]{xcolor}
\usepackage{multirow}
\usepackage{amsmath,amssymb}
\usepackage{adjustbox}
\usepackage{pifont}
\usepackage{tcolorbox}
\usepackage{fontawesome5}
\usepackage{tikz}
\usetikzlibrary{matrix, positioning}

\newcommand{\rev}[1]{\textcolor{black}{#1}}

\newcommand\hl[1]{\textcolor{black}{#1}}

\definecolor{okay}{HTML}{f0eff0}
\definecolor{under}{HTML}{BCC1DE}
\definecolor{over}{HTML}{D1B1BA}

\definecolor{crimson}{HTML}{DC143C}
\definecolor{fogreen}{HTML}{228B22}
\definecolor{daorange}{HTML}{FF8C00}
\definecolor{daviolet}{HTML}{9400D3}
\newcommand{\squa}[1]{\textcolor{#1}{\small$\blacksquare$}}
\newcommand{\staa}[1]{\textcolor{#1}{\small$\bigstar$}}
\newcommand{\diam}[1]{\textcolor{#1}{\small$\blacklozenge$}}
\newcommand{\tria}[1]{\textcolor{#1}{\small$\blacktriangle$}}

\DeclareCaptionStyle{ruled}{labelfont=normalfont,labelsep=colon,strut=off} 
\floatstyle{ruled}
\newfloat{listing}{tb}{lst}{}
\floatname{listing}{Listing}

\title{Human and LLM Biases in Hate Speech Annotations:\\A Socio-Demographic Analysis of Annotators and Targets\thanks{\textcolor{red}{Article published in \textit{ICWSM'25 – 19th AAAI Conference on Web and Social Media}. DOI: https://doi.org/10.1609/icwsm.v19i1.35837. Please, cite the published version.}}}

\author{
    Tommaso Giorgi,\textsuperscript{\rm 2}\textsuperscript{*}
    Lorenzo Cima,\textsuperscript{\rm 1,\rm 2}\thanks{Equal contributions}\thanks{Corresponding author.}
    Tiziano Fagni,\textsuperscript{\rm 1}
    Marco Avvenuti,\textsuperscript{\rm 2}
    Stefano Cresci\textsuperscript{\rm 1}
}
\affiliations{
    \textsuperscript{\rm 1}IIT-CNR, Italy, \\
    \textsuperscript{\rm 2}University of Pisa, Italy \\
    {\{name.surname\}@iit.cnr.it}, tommaso.giorgi@studenti.unipi.it, marco.avvenuti@unipi.it
}

\begin{document}

\maketitle

\begin{abstract}
The rise of online platforms exacerbated the spread of hate speech, demanding scalable and effective detection. However, the accuracy of hate speech detection systems heavily relies on human-labeled data, which is inherently susceptible to biases. While previous work has examined the issue, the interplay between the characteristics of the annotator and those of the target of the hate are still unexplored. We fill this gap by leveraging an extensive dataset with rich socio-demographic information of both annotators and targets, uncovering how human biases manifest in relation to the target's attributes. Our analysis surfaces the presence of widespread biases, which we quantitatively describe and characterize based on their intensity and prevalence, revealing marked differences. Furthermore, we compare human biases with those exhibited by persona-based LLMs. Our findings indicate that while persona-based LLMs do exhibit biases, these differ significantly from those of human annotators. Overall, our work offers new and nuanced results on human biases in hate speech annotations, as well as fresh insights into the design of AI-driven hate speech detection systems.

\noindent {\textbf{Warning:} \textit{This paper contains examples that may be perceived as offensive or upsetting. Reader discretion is advised.}}
\end{abstract}

\section{Introduction}
The proliferation of hateful and toxic speech on online social platforms is among the most severe threats to inclusive online spaces. The exponential increase in user-generated content has made manual hate speech detection and mitigation impractical, driving the development of automated systems~\cite{garg2023handling,cima2024contextualized}. However, the effectiveness of automated systems is heavily contingent upon the quality of their training data, which is typically human-labeled. Challenges arise as hate speech detection is inherently subjective: what one individual may perceive as offensive, another might consider acceptable or even normal. This subjectivity is shaped by the annotators’ backgrounds, experiences, and cultural contexts. As such, socio-demographic attributes such as age, gender, race, and others, can influence how annotators interpret and label content, possibly leading to biases~\cite{davani2023hate}. In turn, biases may seep into the automated detection systems trained on such data, causing severe consequences such as the unwarranted removal of non-hateful content or the disproportionate targeting of certain categories of users~\cite{nogara2024toxic}. 

\begin{figure}[t]
\includegraphics[width=1\columnwidth]{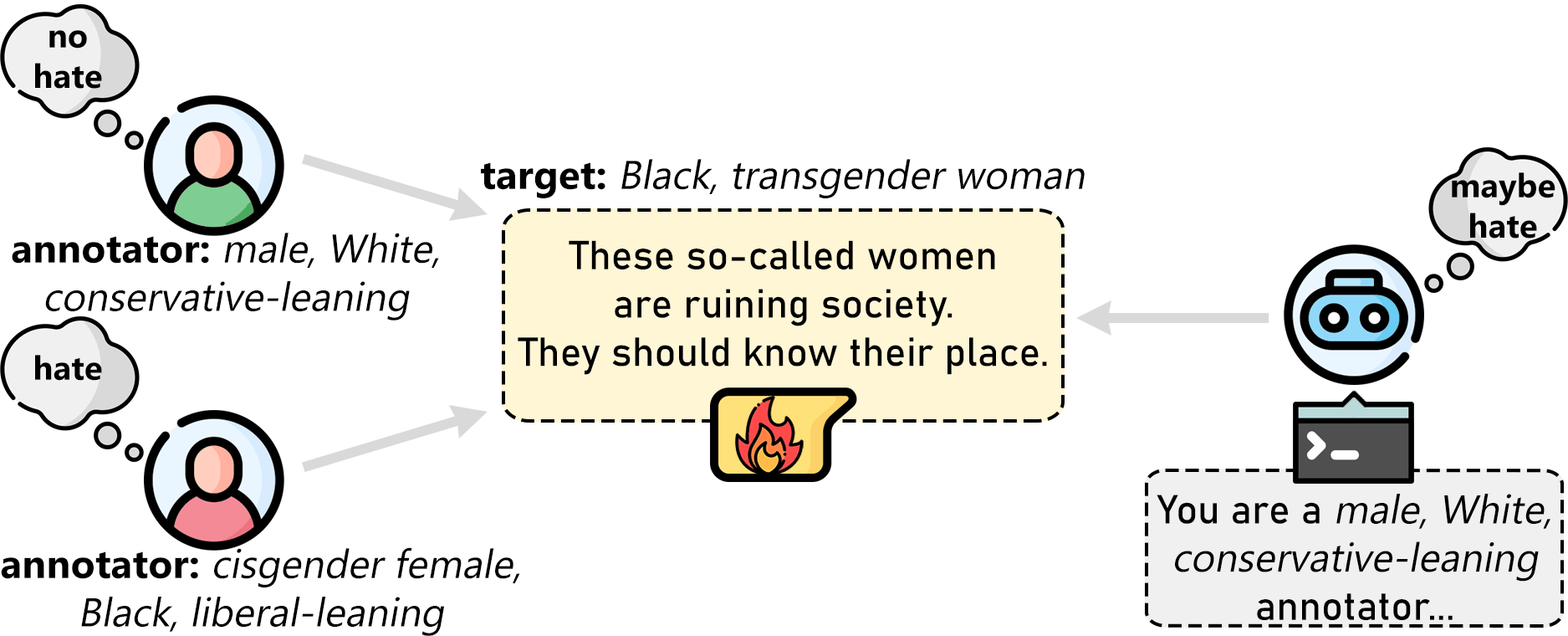}
\caption{Both human and persona-based LLM hate speech annotations can be influenced by the interplay between the \textit{annotator}'s and \textit{target}'s socio-demographic attributes. This possible source of bias is however little-explored.}
\label{fig:problem}
\end{figure}

Large language models (LLMs) have recently extended the set of AI tools in support of platform administrators. However, LLMs are not immune to the biases embedded in the textual data on which they are trained~\cite{baack2024critical}. Consequently, also these models may reflect and perpetuate stereotypes, prejudices, and biases~\cite{perez2023discovering}. Despite the development of various techniques aimed at reducing LLM biases, this remains an outstanding challenge~\cite{ouyang2022training}. Interestingly, the capacity of LLMs to produce subjective outputs can also serve beneficial purposes. LLMs can be personalized and deployed as actors in agent-based simulations of online human interactions~\cite{tornberg2023simulating,park2024generative,rossetti2024social}, which represents a promising way to compensate for the limited availability of platform data in the post-API age. By accurately reproducing certain human biases, persona-based LLMs could lay the groundwork for the next generation of AI-driven social simulations. For these reasons, investigating biases in both human hate speech annotations and persona-based LLMs is a task of great theoretical and practical relevance~\cite{tseng2024two}.

Despite significant efforts, knowledge of human biases in hate speech annotations is still limited. First, many existing studies rely on small datasets. This results in works that only consider a narrow set of socio-demographic attributes, studied in isolation~\cite{prabhakaran2021releasing}. Similar limitations also exist in the nascent literature on LLM biases~\cite{wan2023biasasker}. Most importantly, so far the relationship between the socio-demographic attributes of the \textit{annotator} and those of the \textit{target} of the hate have been overlooked. However, attributes of the target can significantly influence the annotator's perception of what constitutes hate speech. For example, a White, male, conservative-leaning annotator may interpret an offensive post directed at a Black, transgender woman differently than an annotator with a different background, potentially downplaying the harmful impact of the content, as sketched in Figure~\ref{fig:problem}. Understanding such dynamics is essential for accurately mapping human and LLM biases, and for ensuring that hate speech detection is both fair and effective across diverse populations.

\textbf{Research focus.} Motivated by this knowledge gap, we investigate the biases that human annotators and persona-based LLMs exhibit in a hate speech annotation task. We analyze an extensive and publicly available dataset containing 136K hate speech labels assigned by 8K human annotators. For each hateful message, we also analyze the target of the attack. Targets and annotators are described by ten socio-demographic attributes, as summarized in Table~\ref{tab:dataset-small}. Our analysis allows us to answer the following research questions:
\begin{description}
    \item[RQ1a] \textit{Are human annotators more sensitive to hate speech directed at individuals or groups sharing their own socio-demographic attributes?}
    \item[RQ1b] \textit{How do annotators with a specific socio-demographic attribute differ in their labeling of hate speech compared to annotators without that attribute?}
\end{description}
In \textbf{RQ1a}, we hypothesize that identification with the target group (i.e., the annotator's \textit{in-group}) may heighten the annotator's sensitivity to hate, leading them to flag such content more often~\cite{sachdeva2022assessing,hu2024generative}. Conversely, when the target belongs to an \textit{out-group}, the annotator's perception of the hate may be less acute. \textbf{RQ1b} broadens the analysis by examining whether certain groups systematically label content differently than others, in relation to the targets of the hate. This analysis reveals possible disparities in how hate speech is perceived across diverse socio-demographic backgrounds.
\begin{description}
    \item[RQ2a] \textit{How do persona-based LLMs with a specific socio-demographic attribute differ in their labeling of hate speech compared to LLMs without that attribute?}
    \item[RQ2b] \textit{Are persona-based LLM biases the same as those reported by the human annotators they impersonate?} \end{description}
We move our focus from human biases to LLMs'. \textbf{RQ2a} mirrors the analysis in \textbf{RQ1b} by investigating the biases exhibited by persona-based LLMs. Finally, \textbf{RQ2b} determines whether persona-based LLMs replicate the biases observed in human annotators with matching attributes, thereby assessing the extent to which LLMs reflect or diverge from human patterns of bias in hate speech annotation.

 \section{Related Work}
\textbf{Annotator diversity and data quality.} Numerous studies have examined how the characteristics of human annotators impact the quality of training datasets. \citet{geva2019we} demonstrated that datasets generated by a small number of high-quality crowd workers lack diversity, which affects model generalization. Similarly, \citet{wich2020investigating} identified annotator bias in hate speech detection systems and proposed using community detection algorithms to group annotators and mitigate this bias. They also showed that socio-demographic attributes significantly affect bias in hate speech datasets~\cite{al2020identifying}. \citet{parmar2023don} found biases even in the instructions used for crowdsourcing data annotations tasks. Finally, \citet{sap2022annotators} highlighted that annotator identity and beliefs affect toxicity ratings, stressing the need for contextualizing labels with social variables, as done in our work. 

\textbf{Bias in human hate speech annotations.} The impact that biased datasets have on the reliability and fairness of automated hate speech detection systems sparked much research on the issue~\cite{garg2023handling}. \citet{talat2016hateful} showed that cultural and socio-economic backgrounds lead to variability in hate speech annotations. Inherent prejudices further skew labels, as human annotators may be more sensitive to hate speech against familiar groups while underestimating it against less familiar ones. Others found that socio-demographic factors such as age, gender, race and ethnicity, education, and English proficiency, influence annotation outcomes, with younger and minority annotators more likely to label content as hate speech~\cite{al2020identifying,sap2022annotators,davani2023hate}. Additionally, annotator beliefs and demographics can introduce further inconsistencies in crowdsourced annotations~\cite{hettiachchi2023crowd}. Although much research has explored socio-demographic biases in hate speech annotations~\cite{garg2023handling}, most studies focused on a limited set of attributes, primarily race and ethnicity, gender, and age. Moreover, none investigated the interplay between annotator and target attributes.

\begin{table*}[t]
    \setlength{\tabcolsep}{2pt}
    \small
    \centering
    \adjustbox{max width=\textwidth}{
    \begin{tabular}{llrr}
    \toprule
    \textbf{attribute} & \textbf{values} & \textbf{annotators} & \textbf{targets} \\
    \midrule
        age        & \textit{children, teenagers, young adults, middle aged, seniors, other} & 7,907 & 3,522 \\
        disability & \textit{cognitive, hearing impaired, neurological, physical, unspecific, visually impaired, other} & -- & 4,719 \\
        education  & \textit{some high school, high school, some college, college aa, college ba, master, PhD, professional degree} & 7,911 & -- \\
        ideology   & \textit{neutral, slightly liberal, extremely liberal, slightly conservative, conservative, extremely conservative, no opinion} & 7,910 & -- \\
        gender     & \textit{men, non binary, transgender men, transgender unspecified, transgender women, women, other} & 7,950 & 51,325 \\
        income     & \textit{$<$10K, 10K--50K, 50K--100K, 100K--200K, $>$200K} & 7,906 & -- \\
        origin     & \textit{immigrant, migrant worker, specific country, undocumented, other} & -- & 32,503 \\
        race       & \textit{asian, black, latinx, middle eastern, native american, pacific islander, white, other} & 8,589 & 67,625 \\
        religion   & \textit{atheist, buddhist, christian, hindu, jewish, mormon, muslim, other, nothing} & 8,053 & 32,663 \\
        sexuality  & \textit{bisexual, gay, lesbian, straight, other} & 7,886 & 36,521 \\
    \bottomrule  
    \end{tabular}}
    \caption{Dataset overview. Annotators and hate targets are characterized by ten socio-demographic attributes, each assuming multiple values. Appendix Table~\ref{tab:dataset-large} reports the detailed distribution of annotators and targets across attributes and values.}
    \label{tab:dataset-small}
\end{table*}
 
\textbf{Demographic alignment of LLMs.} LLM prompting has been used to experiment with persona-based and role-playing models, to test the alignment of LLM predictions with human opinions~\cite{tseng2024two}. Studies involved replicating famous cognitive and social experiments \cite{aher2023using,srivastava2023beyond}, as well as assessing agreement and personalization capabilities. Among the existing works, \citet{beck2024sensitivity} and~\citet{hu2024quantifying} measured the impact of socio-demographic prompting on model performance, showing that it can enhance zero-shot learning in subjective tasks, \hl{while~\citet{schafer2024demographics} assessed alignment with human annotations.} Additional studies explored the personalities (e.g., Big5, MBTI) exhibited by LLMs when prompted to emulate certain human socio-psychological traits~\cite{rao2023can,la2025open}. Results showed marked differences between models and a moderate predisposition to personalization and psychological prompting. However, \citet{santurkar2023whose} measured the alignment between human and LLM opinions across a wide array of demographic groups and topics, finding substantial misalignment. The misalignment persisted even after explicitly steering the LLMs towards particular groups. Similarly, \citet{lee2023can} investigated the alignment of instruction-tuned LLMs with the disagreement among human annotators, finding that models fail to capture human disagreements. As briefly summarized, there are many contrasting results on the alignment of persona-based LLMs with human opinions. This gap in understanding underscores the need for further research to clarify the extent and conditions under which LLMs mirror human perspectives.

\textbf{Bias in persona-based LLMs.} Multiple recent studies examining social bias in LLMs indicated that the models tend to lean towards left-wing views on issues such as immigration, gun rights, and the political compass test \cite{perez2023discovering,santurkar2023whose}. Additionally, \citet{argyle2023out} showed that GPT-3 can emulate certain human subpopulations, revealing demographically-correlated algorithmic biases; while \citet{simmons2023moral} found political moral biases. It was also shown that persona assignment impacts LLM reasoning and reveals socio-demographic biases, with de-biasing efforts proving largely ineffective \cite{gupta2023bias}. Finally and more akin to our work, \citet{das2024investigating} explored gender, race, religion, and disability biases in hate speech annotations made by GPT-3.5 and GPT-4. Unlike current general research on biases in persona-based LLMs, we specifically focus on a hate speech annotation task, and we assess how LLM biases compare to human biases in this task, for a wide array of socio-demographic attributes.

 \section{Dataset}
The \textsc{Measuring Hate Speech} corpus~\cite{sachdeva2022measuring} is an extensive and publicly available dataset containing 135,556 hate speech labels assigned by 8,472 human annotators to 39,565 distinct social media posts. The posts were collected between March and August 2019 from three major online platforms: Twitter, Reddit, and YouTube. In addition to hate labels, the dataset includes rich socio-demographic information about both the annotators and the targets of the hateful messages. This socio-demographic information spans ten attributes, each with multiple values, resulting in a fine-grained and nuanced resource for in-depth studies of how socio-demographic characteristics influence perceptions of hate. Table~\ref{tab:dataset-small} provides an overview of the dataset, while Appendix Table~\ref{tab:dataset-large} provides detailed information for each attribute and value. The labels were obtained via a data annotation task on Amazon Mechanical Turk, where annotators were asked to classify each post as containing hate (\texttt{hate} label), not containing hate (\texttt{non-hate}), or whether they were uncertain (\texttt{maybe}). On average, each annotator labeled $\sim$17 posts ($\sigma = 3.8$) and each post has been labeled by $\sim$3.5 different annotators ($\sigma = 27.0$). 

As shown in Table~\ref{tab:dataset-small}, annotators are described by eight attributes while hate targets by seven attributes. Five attributes overlap between annotators and targets, while the remaining are only available either for annotators (education, ideology, income) or targets (disability, origin). The distribution of annotators across the available attributes is almost uniform and close to the total number of annotators, meaning that each annotator provided information for the majority of attributes. Appendix Table~\ref{tab:dataset-large} however shows that certain subgroups of annotators are more represented than others. Instead, the distribution of hate targets is very skewed, reflecting the tendency for online hate to be directed towards specific socio-demographic characteristics, such as race and gender. Nonetheless, the availability of a considerable number of annotations also for the least frequent hate targets enables statistically significant analyses across all subgroups.

The dataset follows the FAIR principles. It is available on a prominent cloud storage service,\footnote{\url{https://huggingface.co/datasets/ucberkeley-dlab/measuring-hate-speech} (accessed: 15/12/2024)} making it \textit{findable} and \textit{accessible}. It is \textit{interoperable} in that it is released in a machine-readable format, and \textit{reusable} due to the information contained in the original paper~\cite{sachdeva2022measuring}.  \section{Methodology}

\subsection{Annotation Bias}

\textbf{Definition.} Many definitions of bias have been proposed to date~\cite{garg2023handling}. Informed by these, we define ``annotation bias'' as the systematic deviation in the labeling of hate speech that arises due to the socio-demographic attributes of the annotator or the persona-based LLM. This bias manifests when the annotations of content are influenced not solely by the content itself, but also by the attributes of the annotator, persona-based LLM, or the target, resulting in over- or underestimations of hate speech.

\begin{figure}[t]
\centering
\adjustbox{max width=0.9\columnwidth}{\begin{tikzpicture}
\matrix[matrix of nodes,
    nodes={draw, minimum size=1.15cm, anchor=center, font=\small},
    column sep=-\pgflinewidth, row sep=-\pgflinewidth,
    nodes in empty cells] (m) {
      |[fill=okay]| NN & |[fill=under]| NM & |[fill=under]| NH \\
      |[fill=over]| MN & |[fill=okay]| MM & |[fill=under]| MH \\
      |[fill=over]| HN & |[fill=over]| HM & |[fill=okay]| HH \\
  };

\node[above=1cm of m-1-2] {$A(\mathbf{t} \neq v)$};
  \node[rotate=30, above right=0cm and -0.7cm of m-1-1, font=\small\itshape] {non-hate};
  \node[rotate=30, above right=0cm and -0.7cm of m-1-2, font=\small\itshape] {maybe};
  \node[rotate=30, above right=0cm and -0.7cm of m-1-3, font=\small\itshape] {hate};
  \node[rotate=90, left=1.5cm of m-2-1, anchor=center] {$A(\mathbf{t} = v)$};
  \node[left=0cm of m-1-1, font=\small\itshape] {non-hate};
  \node[left=0cm of m-2-1, font=\small\itshape] {maybe};
  \node[left=0cm of m-3-1, font=\small\itshape] {hate};

\matrix[matrix of nodes, nodes={draw, minimum size=0.5cm}, 
    column sep=0.5cm, row sep=0.5cm, right=0.5cm of m] (legend) {
      |[fill=okay]| \\
      |[fill=over]| \\
      |[fill=under]| \\
  };
  \node[right=0.3cm of legend-1-1, align=left, font=\small] {agreement};
  \node[right=0.3cm of legend-2-1, align=left, font=\small] {$A(\mathbf{t} = v)$\\overestimates};
  \node[right=0.3cm of legend-3-1, align=left, font=\small] {$A(\mathbf{t} = v)$\\underestimates};
\end{tikzpicture}
 }
\caption{Confusion matrix $D$ used to compare labels assigned by annotators with a given socio-demographic attribute $A(\mathbf{t} = v)$ versus those without it $A(\mathbf{t} \neq v)$.}
\label{fig:confusion-matrix}
\end{figure}

\textbf{Operationalization.} Our definition of bias entails the systematic comparison between the labels assigned by human or LLM annotators with certain attributes, with the labels assigned by annotators without such attributes. We operationalize the comparison by extending the method of~\citet{wich2021investigating} as follows. Let $A = \{a_1, \ldots, a_n\}$ be the set of annotators and $G = \{g_1, \ldots, g_m\}$ be the set of targets. We first fix one socio-demographic attribute $\mathbf{t}$ and a specific value of that attribute $\mathbf{t} = v$ (e.g., \textbf{ideology} $=$ liberal). Next, we select all posts labeled by all annotators $A(\mathbf{t} = v)$ with the fixed attribute and value. Depending on the RQ to answer, we further filter the posts by only selecting those whose target $G(\mathbf{t'} = v')$ feature another, possibly different, attribute and value $\mathbf{t'} = v'$. Finally, we compare the labels assigned by the annotators $A(\mathbf{t} = v)$ to those of all other annotators $A(\mathbf{t} \neq v)$ on the same set of selected posts. \hl{This approach to quantifying annotation bias measures differences between the labels assigned by two groups of annotators, without requiring a fixed baseline or a ground-truth. In other words, it captures relative labeling differences between groups without prioritizing one group’s evaluations over another.} 
The comparison is based on the confusion matrix $D$ in Figure~\ref{fig:confusion-matrix}. Rows in $D$ correspond to the labels assigned by the annotators $A(\mathbf{t} = v)$, while columns correspond to the labels assigned by $A(\mathbf{t} \neq v)$. The diagonal represents those instances that are labeled equally by the two groups. The lower triangle represents instances where the annotators $A(\mathbf{t} = v)$ overestimate the hate with respect to $A(\mathbf{t} \neq v)$. Conversely, the upper triangle represents instances where $A(\mathbf{t} = v)$ underestimate hate. We repeat the analysis $\forall \; \mathbf{t},v$. Algorithm \ref{Alg:ConfusionMatrix} in the Appendix provides the pseudo-code implementation of the process used to compute the confusion matrix $D$.

\textbf{Statistical significance.} Our comprehensive approach to investigating bias across a large set of socio-demographic attributes results in a substantial number of comparisons. However, as reported in Appendix Table~\ref{tab:dataset-large}, certain minority groups of annotators and targets are underrepresented in the data. Consequently, some comparisons are based on limited data, and the observed labeling differences may lack statistical significance. \hl{We address this issue by conducting Mann-Whitney tests to assess the statistical significance of the differences between the distributions of labels assigned by any two groups of annotators being compared. Additionally, given the large number of comparisons, we apply the Holm-Bonferroni correction for multiple hypothesis testing. In the remainder, we only report results about those comparisons whose \textit{p}-value $< 0.1$.} Due to data imbalance, we were unable to obtain any significant result for a small set of particularly underrepresented groups, highlighted in gray in Table~\ref{tab:dataset-large}.

\textbf{Measures.} The confusion matrices derived from statistically significant comparisons serve as a foundational tool to identify biases in data~\cite{wich2021investigating}. Then, to highlight the most relevant and severe instances of bias, we propose a set of quantitative indicators to measure the strength, direction, and prevalence of the identified biases:
\begin{description}
    \item[Bias intensity ($I$).] $I$ is obtained as the difference between the sum of values in the lower triangle and the upper triangle of $D$, divided by the sum of both:
    \[
    I = \frac{\sum_{i<j}D_{ij}-\sum_{i>j}D_{ij}}{\sum_{i<j}D_{ij}+\sum_{i>j}D_{ij}}
    \]
    $I \in [-1,+1]$, where $I \approx -1$ indicates a strong tendency to underestimate hate by $A(\mathbf{t} = v)$, while $I \approx +1$ suggests a strong tendency to overestimate. $I \approx 0$ indicates minimal bias. Thus, $I$ conveys both the polarity of the bias (i.e., over- or underestimation) and its strength, offering a clear and concise measure of bias intensity.
    \item[Bias prevalence ($P$).] $P$ quantifies the overall occurrence of bias within the annotation process, irrespective of its intensity. It is calculated as the sum of the values in both the lower and upper triangles of $D$, divided by the sum of all values:
    \[
    P = \frac{\sum_{i<j}D_{ij}+\sum_{i>j}D_{ij}}{\sum_{i,j}D_{ij}}
    \]
    $P \in [0,1]$, where $P \approx 0$ suggests that the bias measured by $I$ occurs in only a minority of cases. Conversely, larger values of $P$ indicate that the bias occurs in a more substantial portion of the annotations, reflecting a non-negligible prevalence. Unlike $I$, which measures the direction and strength of bias, $P$ focuses on the extent to which that bias is prevalent across the entire set of labels.
    \item[Agreement ($\kappa$).] We complement the previous indicators with a standard measure of annotator agreement. Cohen's $\kappa$ gauges the extent of agreement between two groups of annotators, correcting for the agreement that could occur by chance. Here, it contextualizes findings from $I$ and $P$, highlighting whether the observed biases occur in the context of high or low agreement between annotators. \end{description}

\subsection{Large Language Models}

\textbf{Model selection.} We prioritize open-source models over commercial options (e.g., GPT-4) when selecting the LLMs for our experiments. First, open-source models are freely available, which enhances transparency and reproducibility of our research. Additionally, many open-source models have comparable accuracy to that of leading commercial LLMs.\footnote{\url{https://lmarena.ai/?leaderboard} (accessed: 15/12/2024)} Open-source LLMs also require less computational resources, broadening accessibility and easing implementation. These choices make our results generalizable and broadly applicable. Driven by these considerations, we based our model selection on recent results on the performance of open-source LLMs, favoring high-performance and widely-used models~\cite{la2025open}. Our selection includes \textit{Llama-3-8B-Instruct} (8B),\footnote{\url{https://shorturl.at/0xT0b} (accessed: 15/12/2024)} \textit{Phi-3-mini-4k-instruct} (3.8B),\footnote{\url{https://shorturl.at/tOs9i} (accessed: 15/12/2024)} \textit{SOLAR-10.7B-Instruct-v1.0} (10B),\footnote{\url{https://shorturl.at/udLQZ} (accessed: 15/12/2024)} and \textit{Starling-LM-7B-alpha} (7B).\footnote{\url{https://shorturl.at/kqqLq} (accessed: 15/12/2024)}

\textbf{Role-playing.} We adopt a traditional prompt engineering approach to experiment with role-playing LLMs~\cite{tseng2024two}. The prompt includes detailed instructions on how to carry out the hate speech annotation task. Additionally, it dynamically specifies a set of socio-demographic attributes, allowing LLMs to impersonate annotators with certain given attributes. Initially, we experimented with 100 randomly selected posts to manually refine and evaluate various prompts. Through iterative prompting and evaluation, we assessed the quality of the LLMs' output. The final version of the prompt is reported in Appendix Table~\ref{tab:llm-prompt}, together with some alternative versions that we evaluated.

\textbf{Model evaluation.} We assess the suitability of persona-based LLMs to act as hate speech annotators based on two criteria: (\textit{i}) the model's ability to accurately perform hate speech detection, and (\textit{ii}) its sensitivity to personalization, meaning the extent to which the LLM's output changes when personalized with specific socio-demographic attributes. To evaluate the first criterion, we used the base versions (i.e., neither personalized nor fine-tuned) of each model to label the posts in our dataset. We then compared their performance against that of human annotators to identify which base models excel at hate speech detection. For the second criterion, we assessed sensitivity to personalization by focusing on the 11,810 posts in our dataset (30\%) where human annotators disagreed about the assigned label. We personalized each LLM with socio-demographic attributes and tasked them with labeling this subset of posts. We then measured the agreement $\kappa$ between the personalized outputs and those of the base, non-personalized versions of each model. This analysis reveals which LLMs are most affected by the personalization process---indicated by lower $\kappa$---thus identifying the model for which personalization is most effective.  \section{Analyses and Results}

\begin{figure}
\centering
\includegraphics[width=0.85\columnwidth]{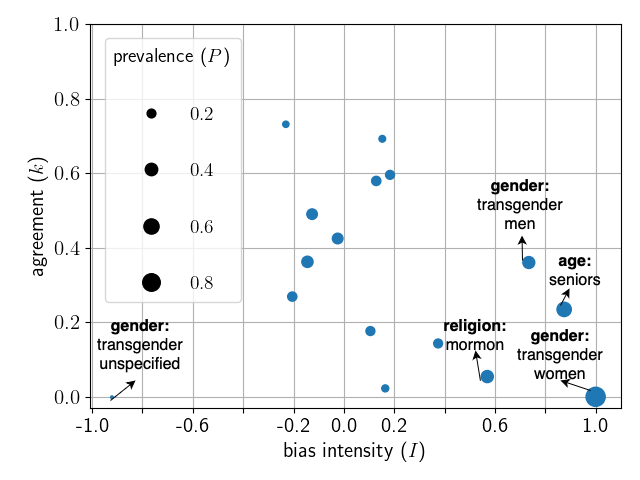}
\caption{\hl{Sensitivity to in-group hate. Points correspond to the statistically significant biases that we found. Most biases have mild intensity and low prevalence, with a few remarkable exceptions highlighted in the plot.}}
\label{fig:scatterplot-human-bias-in-group}
\end{figure}

\subsection{RQ1a: Sensitivity to In-Group Hate}
\textbf{Analysis.} We assess whether annotators are more sensible to hate directed at their own in-group. We iteratively fix an attribute and value. Then, we perform the analysis by selecting all labels assigned by annotators with the fixed attribute and value to posts targeted at the same group, and by comparing these annotations to those given to the same posts by annotators without the fixed attribute and value.

\begin{figure*}
\includegraphics[width=\textwidth]{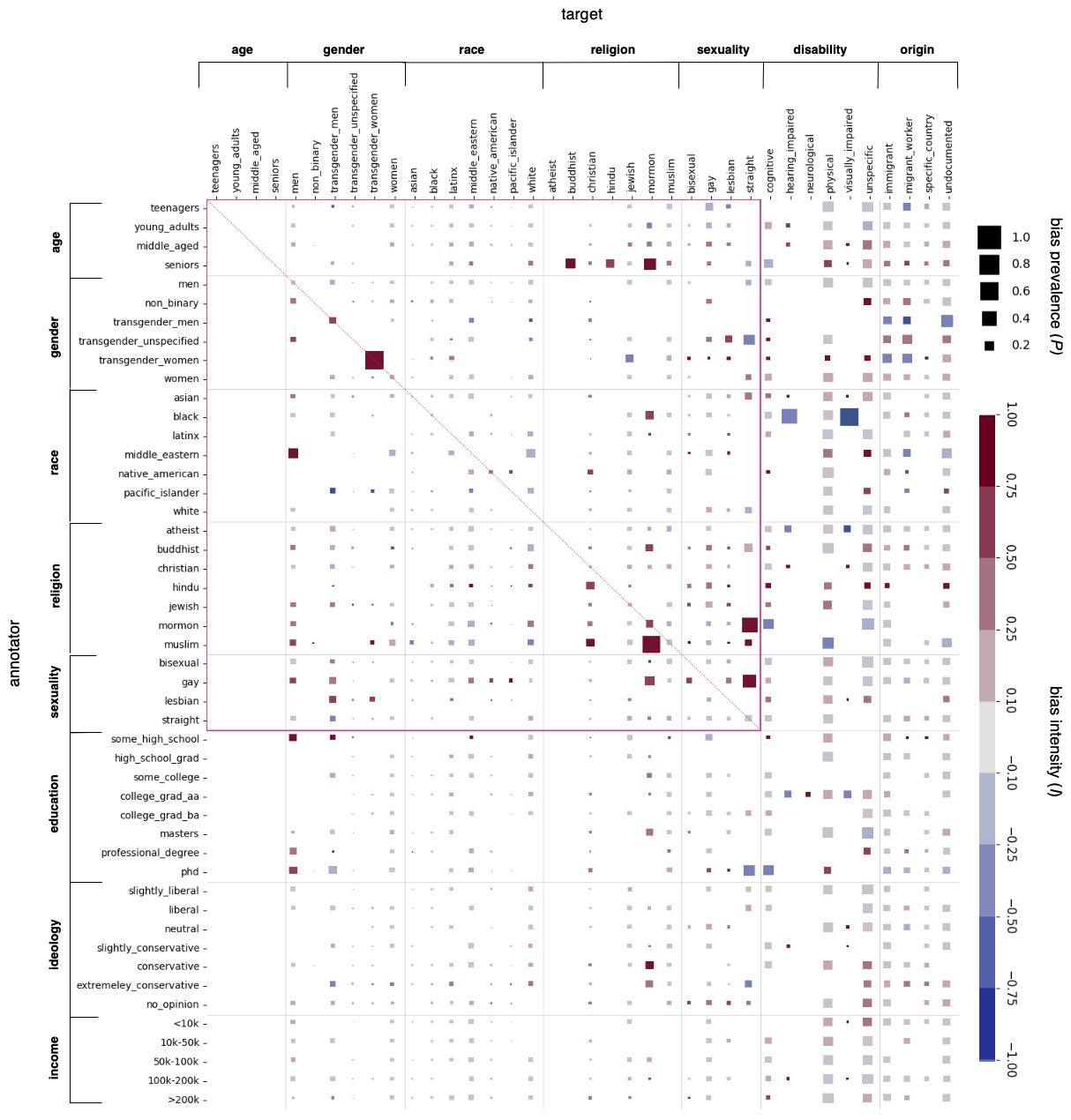}
\caption{\rev{Intensity ($I$) and prevalence ($P$) of human biases in hate speech annotations, for all possible combinations of annotator (\textit{y} axis) and target (\textit{x} axis) socio-demographic attributes. Cell color corresponds to intensity while cell area is proportional to prevalence. Red-colored cells mark statistically significant hate overestimations, while blue-colored ones denote significant underestimations. The magenta-colored top-left square corresponds to the socio-demographic attributes that are available for both the annotators and targets (age, gender, race, religion, sexuality), as thoroughly reported in Appendix Table~\ref{tab:dataset-large}. The dashed diagonal corresponds to in-group annotations (i.e., same annotator and target attribute). Attributes outside of the top-left square are only available either for annotators (education, ideology, income) or targets (disability, origin). Annotator and target attributes for which we did not obtain any significant result (e.g., \textbf{ideology}: \textit{extremely liberal}) are not shown.
To ease the exploration of our results, an interactive version of this visualization by~\citet{Abrate2024} is available at \url{https://webvis.github.io/annotator-target-hate-speech-biases/}}.}

\label{fig:heatmap-human-bias-intensity}
\end{figure*}

\begin{figure*}
    \centering
    \includegraphics[width=0.45\textwidth]{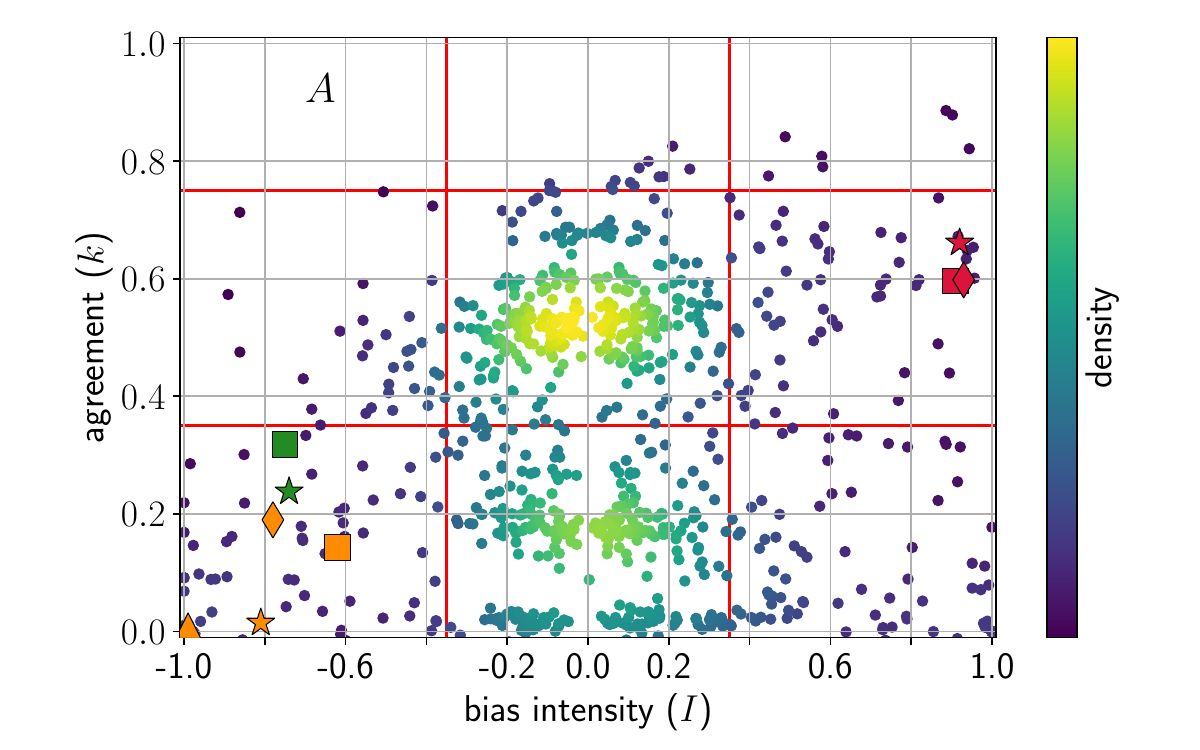}\hspace{0.05\textwidth}\includegraphics[width=0.45\textwidth]{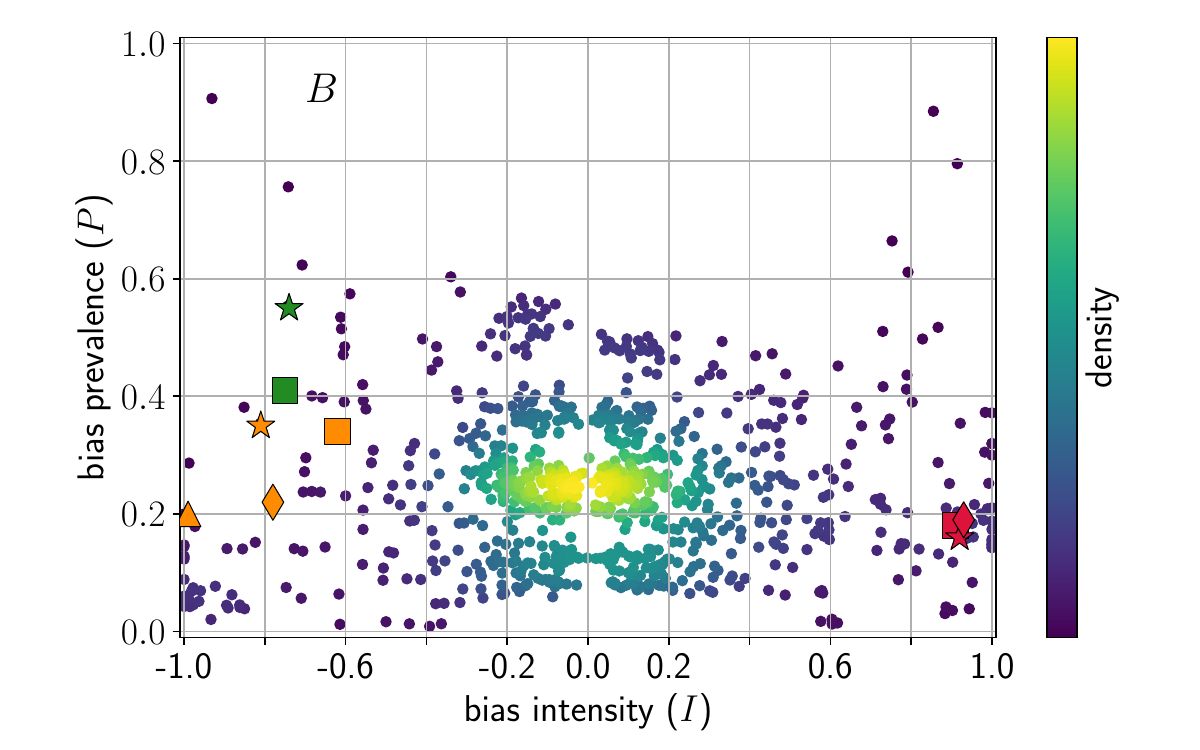}\caption{\rev{Distribution of statistically significant human biases in the intensity-agreement (panel \textit{A}) and intensity-prevalence (panel \textit{B}) space. Red lines in panel \textit{A} delimit different regions of the space. Panel \textit{A} shows the existence of biases with strong intensity and low agreement, while panel \textit{B} highlights some biases characterized by both large prevalence and intensity. Some examples are highlighted with colored symbols and further described in Table~\ref{tab:human-bias-examples}.}}
    \label{fig:scatterplot-human-vs-human}
\end{figure*}

\textbf{Results.} \rev{We found statistically significant labeling differences for 16 of 29 (55\%) socio-demographic attributes. For each of these cases, we measured bias intensity ($I$), prevalence ($P$), and annotator agreement ($\kappa$). Results are shown in Figure~\ref{fig:scatterplot-human-bias-in-group}. Regarding the 16 significant differences, 10 (62\%) were overestimations ($I > 0$) and 6 were underestimations. Bias intensity was generally mild or moderate, with \(I \leq \pm 0.25\) in most cases.} However, we identified strong biases (\(I > \pm 0.5\)) in some instances. As expected, the strongest biases occur when there is little agreement ($\kappa < 0.4$) between the annotator groups. Regarding prevalence $P$, some biases are overall marginal since they occur very infrequently. There are situations, however, for which we found biases that are both intense and with large prevalence ($P > 0.4$), such as those related to \textit{transgender} annotators and targets. Other cases with considerable biases are those annotators and targets in the \textit{seniors} age group and of \textit{Mormon} religion, when compared to other annotators.

\textbf{Insights.} These results indicate that while in-group hypersensitivity is not universal, there is a mild tendency to overestimate in-group hate with respect to out-group annotators. Moreover, significant and sometimes substantial biases are present in specific socio-demographic contexts.

\subsection{RQ1b: Human Biases in Annotator-Target Dynamics}
\textbf{Analysis.} We broaden our analysis of annotation bias in hate speech by examining all combinations of annotator and target attributes. For each iteration, we fix attribute $\mathbf{t} = v$ for annotators and attribute $\mathbf{t'} = v'$ for hate targets. We then compare the labels assigned by annotators with $\mathbf{t} = v$ to posts directed at targets with $\mathbf{t'} = v'$, against the labels provided by annotators with $\mathbf{t} \neq v$. This approach allows us to systematically assess how the interplay between annotator and target attributes influence hate speech labeling.

\textbf{Results.} Figure~\ref{fig:heatmap-human-bias-intensity} shows a heatmap of bias intensity ($I$) and prevalence ($P$) for the explored combinations of annotator and target attributes. Intensity is color-coded while prevalence is proportional to cell area. Statistical significance is assessed at the 0.1 level with a Mann-Whitney test with Holm-Bonferroni correction. Missing cells denote non-significant results. The figure caption provides further information on the heatmap and its interpretation.

Figure~\ref{fig:heatmap-human-bias-intensity} provides nuanced results on hate speech annotation biases. When observed row-wise, it allows identifying groups of annotators with the tendency to systematically over- or underestimate hate. As a notable example, our analysis reveals a marked age bias: younger annotators (\textit{teenagers} and \textit{young adults}) generally tend to underestimate hate, while older ones (\textit{middle aged} and \textit{seniors}) overestimate. This bias might be explained by the familiarity that younger annotators have with the toxic language of social media, which might desensitise them to such expressions, contrarily to older annotators who might be less familiar with it. Other groups who overestimate hate speech compared to other annotators are those with a \textit{high school} degree; \textit{religious} people; \textit{transgender women}; and those who identify as \textit{gay}. Conversely, among those who underestimate the most are \textit{atheists}; those with a \textit{PhD} degree; \textit{men}, \textit{transgender men}, and \textit{non-binary} people; and those who identify as \textit{lesbian}. These results indicate possible education, religion, gender, and sexual orientation biases. When observed column-wise, Figure~\ref{fig:heatmap-human-bias-intensity} allows identifying the targets of hate that are overall flagged more or less frequently. For example, human annotators show heightened sensitivity to offenses based on \textit{disability} and \textit{sexual orientation}. Instead, hate directed at certain minorities, such as \textit{migrant workers} and \textit{transgender} individuals, tends to be underestimated. 

\begin{table}[t]
    \small
    \setlength{\tabcolsep}{2pt}
    \centering
    \adjustbox{max width=\columnwidth}{
    \begin{tabular}{cllrrrr}
    \toprule
    &&&& \multicolumn{2}{c}{\textbf{bias}} \\
    \cmidrule{5-6}
    & \textbf{annotator} & \textbf{target} & \multicolumn{1}{c}{\#} & \multicolumn{1}{c}{$I$} & \multicolumn{1}{c}{$P$} & \multicolumn{1}{c}{$\kappa$} \\
    \midrule
    \squa{crimson} & g: transgender women& s: bisexual         & 511 & 0.91 & 0.18 & 0.596 \\
    \staa{crimson} & g: transgender women& s: gay             & 4,880 & 0.92 & 0.16 & 0.661 \\
    \diam{crimson} & g: transgender women& s: lesbian           & 639 & 0.93 & 0.19 & 0.598 \\
    \midrule
    \squa{daorange} & r: atheist          & d: hearing impaired     & 183 & -0.62 & 0.34 & 0.143 \\
    \staa{daorange} & r: atheist          & d: visually impaired    & 209 & -0.81 & 0.35 & 0.015 \\
    \diam{daorange} & a: young adults     & d: hearing impaired     & 249 & -0.78 & 0.22 & 0.190 \\
    \tria{daorange} & a: young adults     & d: visually impaired    & 240 & -1.00 & 0.20 & 0.000 \\
    \midrule
    \squa{fogreen} & g: non binary       & s: straight         & 195 & -0.75 & 0.41 & 0.318 \\
    \staa{fogreen} & g: transgender unspecified & s: straight  & 141 & -0.74 & 0.55 & 0.238 \\ 
    \bottomrule
    \multicolumn{7}{p{\dimexpr\linewidth-2\tabcolsep}}{\footnotesize g:~gender; s:~sexuality; r:~religion; a:~age; d:~disability} \\
    \end{tabular}}
    \caption{Examples of human biases. For each example we report the number of considered labels (\#), and the values of bias intensity ($I$), prevalence ($P$), and agreement ($\kappa$). \hl{All reported examples are statistically significant at the 0.01 level, as confirmed by a Mann-Whitney test with Holm-Bonferroni correction.} Colored symbols map examples into the spaces of Figure~\ref{fig:scatterplot-human-vs-human}.}
    \label{tab:human-bias-examples}
\end{table}
 
The analysis of Figure~\ref{fig:heatmap-human-bias-intensity} reveals general biases either affecting groups of annotators or targets. Now we dig deeper by considering the interplay between annotator and target attributes. Figure~\ref{fig:scatterplot-human-vs-human} shows scatterplots where each statistically significant case previously identified is mapped in the $I$-$\kappa$ and $I$-$P$ spaces. Red lines in Figure~\ref{fig:scatterplot-human-vs-human}\textit{A} delimit different regions of the $I$-$\kappa$ space. The majority of points lay in the central region, corresponding to cases of moderate to strong agreement between the groups of annotators (\(0.35 \leq \kappa \leq 0.75\)) and to low bias intensity (\(I \leq \pm 0.35\)). Another dense area of the plot corresponds to low agreement (\(\kappa \leq 0.35\)) and low bias intensity. These points indicate situations where the compared groups of annotators frequently disagreed on labels, yet these disagreements balanced out without resulting in a marked trend of over- or underestimation. This finding underscores the distinction between agreement and bias, demonstrating that low annotator agreement does not necessarily lead to bias. Contrarily, the bottom left and right corners of Figure~\ref{fig:scatterplot-human-vs-human}\textit{A} correspond to low agreement and strong bias intensity. Table~\ref{tab:human-bias-examples} highlights some of the observed biases that arise from interesting combinations of annotator and target attributes. For example, \textit{transgender women} showed heightened sensitivity to hate directed at \textit{bisexuals}, \textit{gays}, and \textit{lesbians}. Conversely, \textit{atheists} and \textit{young adults} underestimated hate towards \textit{hearing} and \textit{visually impaired} individuals, while \textit{transgender unspecified} and \textit{non-binary} people underestimated hate towards individuals with \textit{straight} sexual orientation. Finally, the top-right corner of Figure~\ref{fig:scatterplot-human-vs-human}\textit{A} corresponds to strong agreement (\(\kappa > 0.75\)) but large bias intensity (\(I > 0.35\)). These cases are characterized by low bias prevalence ($P < 0.42$), which is shown in Figure~\ref{fig:scatterplot-human-vs-human}\textit{B}, suggesting that the observed biases occur infrequently. The strong agreement underscores that these biases are not widespread, making them less relevant overall.

\textbf{Insights.} Our results reveal multiple biases of the annotators and the targets independently, as well as others that arise from specific combinations of annotator and target attributes. These findings show that different annotator groups react distinctly depending on the socio-demographic characteristics of the hate target. While some of the identified biases are mild, others are pronounced both in terms of intensity and prevalence, highlighting that both the individual attributes of annotators and targets, and their interaction, can significantly influence hate speech labeling decisions. 

\begin{table}[t]
    \footnotesize
    \setlength{\tabcolsep}{2pt}
    \centering
    \adjustbox{max width=\columnwidth}{
    \begin{tabular}{lcrrrcrr}
    \toprule
        && \multicolumn{3}{c}{\textbf{hate speech detection}} && \multicolumn{2}{c}{\textbf{personalization}} \\
        \cmidrule{3-5}\cmidrule{7-8}
        \textbf{model} && \textit{macro F1 $\uparrow$} & \textit{weighted F1 $\uparrow$} & \textit{accuracy $\uparrow$} && \multicolumn{2}{r}{\textit{agreement ($\kappa$) $\downarrow$}} \\
        \midrule
        \textit{Llama3}     && 0.40 & 0.56 & 0.59 && \multicolumn{2}{r}{0.88} \\
        \textit{Phi3}       && 0.43 & 0.57 & 0.57 && \multicolumn{2}{r}{\underline{0.79}} \\
        \textit{Solar}      && \textbf{0.53} & \underline{0.69} & \underline{0.66} && \multicolumn{2}{r}{\textbf{0.72}} \\
        \textit{Starling}   && \underline{0.52} & \textbf{0.73} & \textbf{0.75} && \multicolumn{2}{r}{0.91} \\
        \bottomrule  
    \end{tabular}}
    \caption{LLMs evaluation results in terms of the base models' capacity to detect hateful posts and their sensitivity to personalization. Best results in each evaluation metric are shown in \textbf{bold}, second-bests are \underline{underlined}.} \label{tab:models}
\end{table}
 
\subsection{RQ2a: LLM Biases in Annotator-Target Dynamics}
\textbf{Best model selection.} We investigate the biases of persona-based LLMs, similarly to what we did in \textbf{RQ1b} for human annotators. Although we experimented with the four LLMs in Table~\ref{tab:models}, due to space constraints we only present results of the most effective and relevant one. Table~\ref{tab:models} reports the performance of the base LLMs in a hate speech detection task, and their sensitivity to personalization. The former is measured with standard machine learning evaluation metrics, while greater sensitivity to personalization is reflected by lower agreement (i.e., more diversification) with the non-personalized version of the same model. In the remainder we present results from \textit{Solar}, given that it achieved overall second-best performance in hate speech detection and has the best sensitivity to personalization. 

\textbf{Analysis.} We instruct \textit{Solar} to impersonate human annotators with specific socio-demographic attributes. For each annotation in the dataset, we personalize the LLM with the exact socio-demographic attributes of the corresponding human annotator and ask it to label the same post that the human had labeled. We repeat this process for all 136K annotations, creating an LLM-labeled copy of the human-labeled dataset. We then iteratively fix an attribute \(\mathbf{t} = v\) for the LLM and another attribute \(\mathbf{t'} = v'\) for the hate targets. Finally, we compare the labels given by the persona-based LLM with \(\mathbf{t} = v\) to posts directed at targets with \(\mathbf{t'} = v'\), against the labels given by the persona-based LLM without \(\mathbf{t} = v\), thus investigating the impact of LLM socio-demographic attributes on labeling behavior.

\begin{figure}
    \includegraphics[width=\columnwidth]{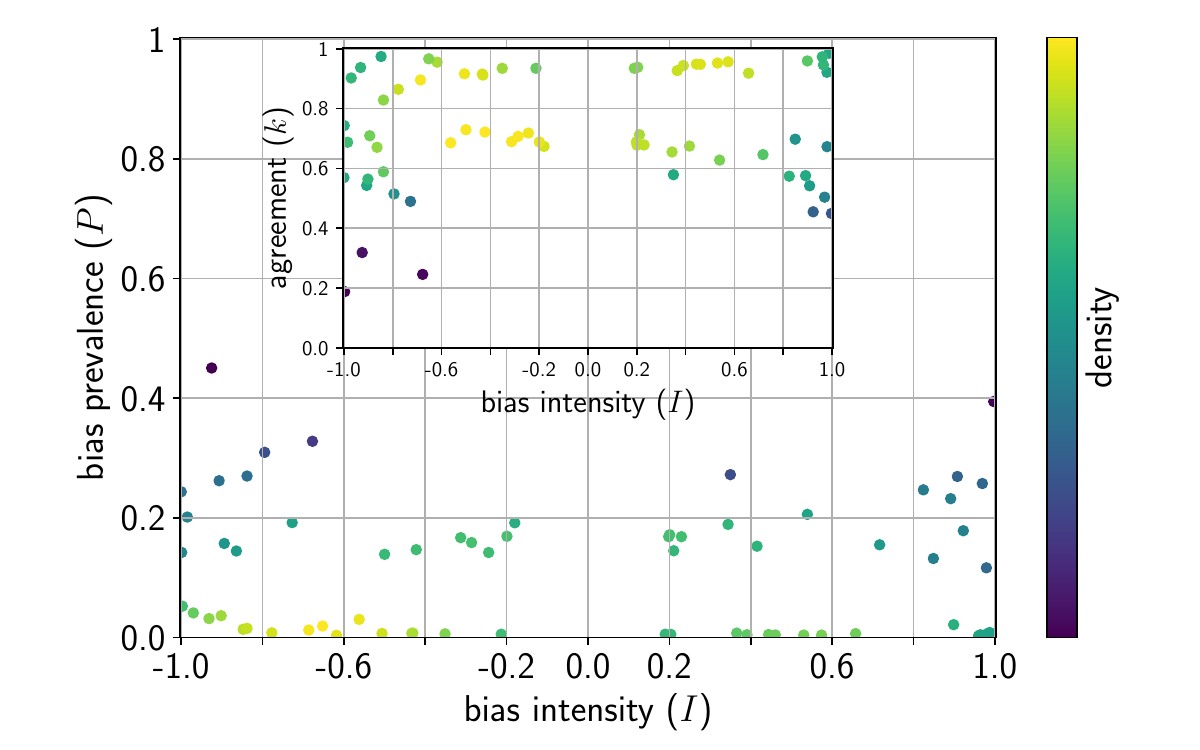}
    \caption{\rev{Distribution of statistically significant LLM biases in the intensity-agreement (inset) and intensity-prevalence (outer figure) space. Differently from humans, persona-based LLMs exhibit strong agreement. No biases exhibit both high intensity and prevalence, but some display high intensity with notable prevalence.}}
    \label{fig:scatterplot-llm-vs-llm}
\end{figure}

\begin{figure*}
    \centering \includegraphics[width=0.33\textwidth]{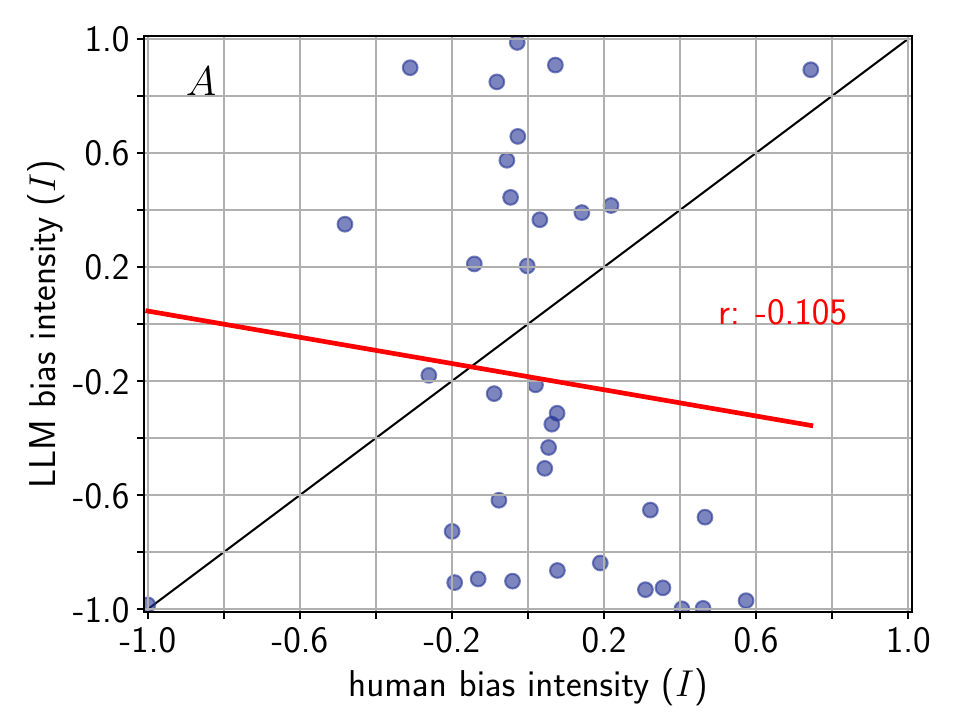}\includegraphics[width=0.33\textwidth]{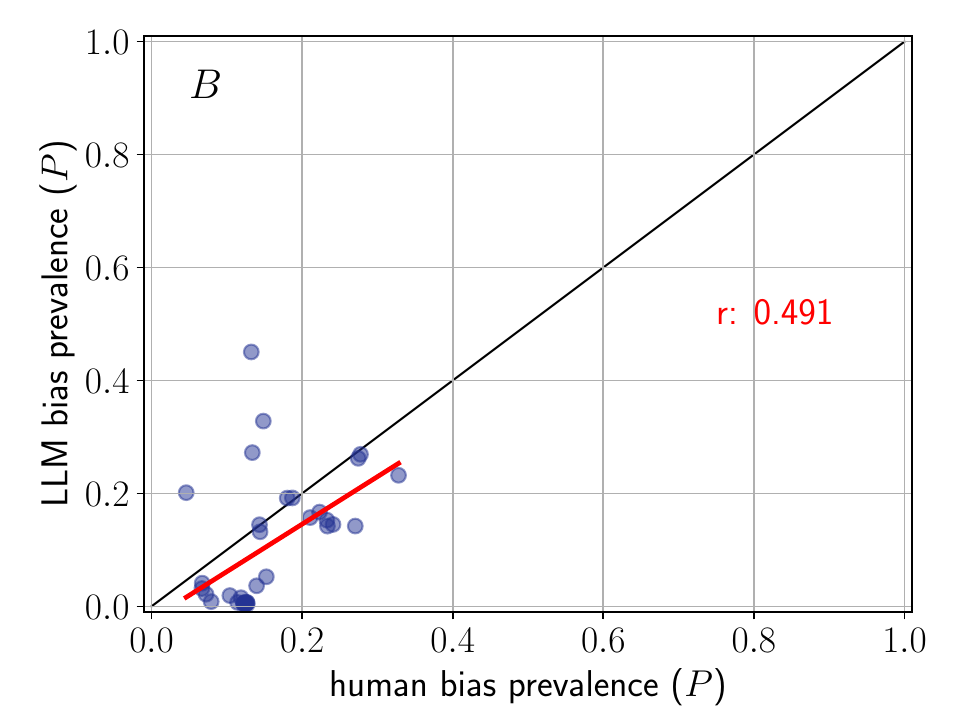}\includegraphics[width=0.33\textwidth]{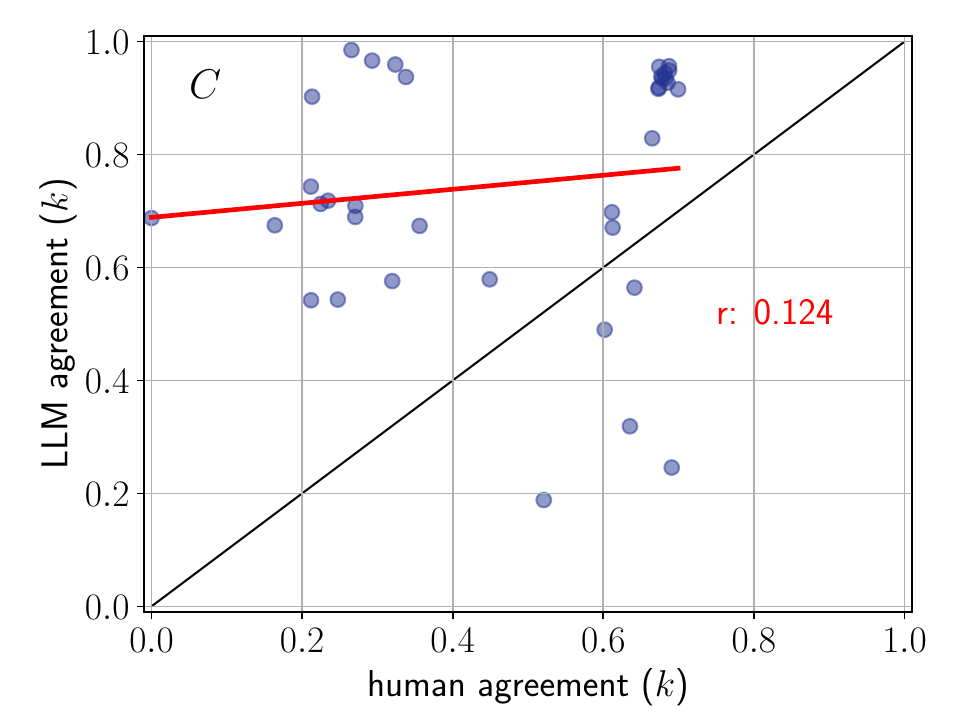}\caption{\rev{Comparison of human and LLM biases across bias intensity (panel \textit{A}), prevalence (panel \textit{B}), and agreement (panel \textit{C}). Each point represents a combination of annotator and target attributes where both humans and persona-based LLMs exhibited a statistically significant bias. Red lines and annotations report Pearson correlations. The intensity (i.e., strength and polarity) of LLM biases is uncorrelated with that of humans. However, we measured a moderate positive correlation in prevalence.}}
    \label{fig:scatterplot-llm-vs-human}
\end{figure*}

\textbf{Results.} \rev{We observed statistically significant labeling differences in 65 out of 1,950 (3.33\%) combinations of annotator and target attributes when using persona-based LLMs. In comparison, human-generated annotations showed significant differences in 900 out of 1,950 (46\%) combinations.} The relatively low number of significant results for persona-based LLMs, especially compared to human annotators, suggests that the personalization had a limited impact on the hate speech labels assigned by the models and that the socio-demographic attributes embedded in the LLMs did not lead to substantial labeling differences. Figure~\ref{fig:scatterplot-llm-vs-llm} presents each significant combination in the \(I\)-\(\kappa\) space (inset) and \(I\)-\(P\) space (outer figure). Unlike human annotators, persona-based LLMs largely exhibited substantial agreement (\(\kappa \geq 0.6\)) in their label assignments. In terms of bias intensity, we identified a range from mild to severe, with both notable over- and underestimations. However, the high level of agreement resulted in relatively low bias prevalence, indicating that the majority of detected biases were infrequent. Nevertheless, we identified several biases characterized by both high intensity and substantial prevalence. For instance, LLMs impersonating annotators who identified as \textit{Christian} consistently underreported hate directed at \textit{bisexual} individuals (\(I = -0.84\), \(P = 0.27\), \(p < 0.01\)). Similarly, LLMs impersonating annotators with \textit{straight} sexual orientation underreported hate targeted at \textit{men} (\(I = -0.91\), \(P = 0.26\), \(p < 0.01\)), while those impersonating \textit{bisexual} (\(I = 0.91\), \(P = 0.27\), \(p < 0.01\)) and \textit{gay} (\(I = 0.89\), \(P = 0.23\), \(p < 0.01\)) annotators overreported it.

\textbf{Robustness.} We experimented with alternative prompts to assess whether prompt variations would yield better LLM personalization. Following~\citet{beck2024sensitivity} we tested additional prompts by reducing the number of socio-demographic attributes passed to the model, and by simplifying the instructions for the annotation task. These alternative prompts are reported in Appendix Tables~\ref{tab:llm-prompt-2} and~\ref{tab:llm-prompt-3}. However, \textit{Solar}'s sensitivity to personalization decreased with both alternative prompts, respectively with $\kappa = 0.81$ and $\kappa = 0.82$, instead of $\kappa = 0.72$ (lower is better).

\textbf{Insights.} Our findings suggest that persona-based LLMs exhibit fewer labeling differences and, consequently, fewer biases (according to our definition) than human annotators. The reduced variability in the labels assigned by these LLMs, regardless of the socio-demographic attributes they were prompted to emulate, indicates that the model's personalization may not fully capture the nuanced biases present in human behavior. Despite this, we identified several marked biases against minority groups, which could negatively impact the performance of automatic hate speech detection. 

\subsection{RQ2b: Comparison of Human and LLM Biases}
\textbf{Analysis.} We assess the extent to which the biases exhibited by persona-based LLMs are similar to those of human annotators. Concretely, we focus on the statistically significant cases in which both humans and persona-based LLMs exhibit a bias for the same combination of annotator and target attributes. For each of these cases, we compare bias intensity, bias prevalence, and agreement. This approach allows us to measure the correlation between the considered bias indicators of humans and LLMs, providing insights into how closely persona-based LLMs replicate human biases.

\textbf{Results.} \rev{We found 36 statistically significant ($p < 0.1$) cases in which human annotators and persona-based LLMs exhibited a bias for the same combination of annotator and target attributes. As shown in Figure~\ref{fig:scatterplot-llm-vs-human}\textit{A}, the Pearson correlation $r = -0.105$ between the \textit{intensity} of human and LLM biases suggests the lack of a linear relationship. This near-zero correlation indicates that, overall, the biases exhibited by persona-based LLMs do not correspond to those observed in human annotators. A notable exception is the marked tendency that \textit{men} annotators---both human and LLM-impersonated---exhibit to overestimate hate towards \textit{gay} individuals, with bias intensity $I = 0.74$ and $I = 0.89$ for humans and LLMs respectively. Instead, Figure~\ref{fig:scatterplot-llm-vs-human}\textit{B} shows that the correlation between the \textit{prevalence} of human and LLM biases is moderately strong, with $r = 0.491$. This suggests that the biases that are (in)frequent among human annotators tend to be relatively (in)frequent among persona-based LLM annotations as well. Finally, the correlation $r = 0.124$ between the \textit{agreement} measured for human and LLM biases, shown in Figure~\ref{fig:scatterplot-llm-vs-human}\textit{C}, indicates a weak positive relationship. This reflects a slight tendency for the level of agreement among human annotators (with and without a certain attribute) to correspond with the level of agreement among LLMs (with and without the same attribute).} However, the weak correlation suggests that while there is some degree of similarity in how humans and LLMs agree within and across certain socio-demographic attributes, the factors influencing agreement among humans are not fully captured by LLMs. As already observed in Figure~\ref{fig:scatterplot-llm-vs-llm}, this is mainly due to the large agreement exhibited by the persona-based LLMs, which is indicative of strong labeling consistency even when LLMs are personalized with different attributes. 

\textbf{Robustness.} Results obtained with \textit{Llama3}, \textit{Phi3}, and \textit{Starling} qualitatively confirm those of \textit{Solar}, with correlations between human and LLM bias intensity ranging from $r = 0.108$ (\textit{Llama3}) to $r = -0.137$ (\textit{Phi3}). These additional figures show invariance of our results to the specific LLM used and reinforce the conclusion that persona-based LLMs have limited capacity to reproduce human biases. 

\textbf{Insights.} The persona-based LLMs that we used were, in general, unable to replicate human biases. This was particularly evident with respect to the strength and polarity of the biases. However, biases exhibited by persona-based LLMs occurred with a moderately similar frequency than the corresponding human biases, highlighting some similarities between the two. These results suggest that, although persona-based LLMs can mimic some aspects of human bias, they do not closely replicate the nuanced biases present in human hate speech annotations. 
 \section{Discussion and Conclusions}
Our results reveal that human annotators exhibit a mild tendency to overestimate in-group hate (\textbf{RQ1a}), and that different socio-demographic attributes among annotators lead to diverse labeling patterns (\textbf{RQ1b}). Furthermore, LLMs show little sensitivity to personalization via prompting, which likely accounts for the few biases observed, though they still display significant biases against certain minority groups (\textbf{RQ2a}). However, persona-based LLMs do not fully replicate human biases, demonstrating minimal alignment with human annotators (\textbf{RQ2b}).

\textbf{Bias intensity and prevalence.} Our methodology introduces two bias indicators---intensity ($I$) and prevalence ($P$)---that measure the severity and frequency of annotation biases. While $P$ indicates the frequency of disagreement, $I$ measures the extent of imbalance when such disagreement occurs. Clearly, the most concerning biases are those with both high intensity and prevalence. \hl{However, even biases with high intensity and small but non-negligible prevalence are attention worthy. These correspond to situations where disagreement between groups of annotators occurs overall infrequently, as represented by the low prevalence. However, when disagreements do occur, one group systematically over- or underestimates hate, as reflected by the large intensity. These cases may also warrant further investigation.} 

\textbf{Human biases.} \hl{We systematically applied our methodology to identify and characterize human over- and underestimations in hate speech annotations, uncovering a large number of biases with varying intensity and prevalence. We remark however that we lack a reference or ground-truth value for ``correct'' hate speech annotations. Therefore, in general, our results should be interpreted as systematic and statistically significant differences between the labels assigned by groups of annotators with different socio-demographic attributes. In other words, while some of the identified labeling differences correspond to negative biases to correct, other instances---such as some overestimations by in-group annotators---might be indicative of genuine insights that annotators with a different background might have missed.}
To this regard, our study favors breadth over depth, and it is beyond our scope to investigate the identified biases in detail. However, our work provides a foundation for domain experts to conduct in-depth studies on specific biases, exploring their origins and potential solutions, if needed. To foster this line of work, we publicly release the complete list of statistically significant biases that we found, and the indicators that characterize them.\footnote{\url{https://doi.org/10.5281/zenodo.13906773}}

\textbf{LLM biases.} We also investigated the biases exhibited by persona-based LLMs, and compared them to human ones. Current research in this area is facing a significant tension: \hl{on one hand, we strive for unbiased LLMs to avoid propagating human prejudices~\cite{perez2023discovering}, especially when these LLMs act as arbiters in content moderation decisions.} On the other hand, there is a compelling case for developing LLMs that faithfully reproduce human behaviors---including biases---as this would open new frontiers in social simulations~\cite{tornberg2023simulating,park2024generative,rossetti2024social}. Our results in this regard are concerning. First, we found that persona-based LLMs obtained via prompting do exhibit biases. More troubling, however, is the finding that there is very little alignment between the biases exhibited by LLMs and those exhibited by humans. This suggests that current persona-based LLMs may fail to accurately reflect the complex and nuanced biases present in human behavior. This raises concerns about the reliability of using prompting techniques to create persona-based LLMs for social simulations and underscores the need for further research into the sources of LLM biases.

\textbf{LLM personalization.} Our study is the first to analyze bias in hate speech annotations at such a fine-grained level. Thus, our results about the inability of persona-based LLMs to reproduce human biases might stem from two factors: the LLMs' inability to replicate fine-grained biases, or the ineffectiveness of the personalization process. About the former, more powerful LLMs than those tested herein could potentially overcome the issue, resulting in more accurate human representations. Instead, regarding personalization, we found a limited capacity of the LLMs to modify their behavior and adapt to specific personas. This result reinforces the current body of work that questions the ability of LLMs to faithfully reproduce human behaviors~\cite{santurkar2023whose,lee2023can}. \hl{Nonetheless, although we used a widely adopted state-of-the-art approach for personalization~\cite{beck2024sensitivity}, different personalization methods might yield better results. Among them are approaches based on fine-tuning rather than prompting~\cite{agiza2024politune}, or those relying on specific token sampling functions to induce controlled bias in LLM answers.}

\textbf{Implications.} In addition to being a valuable resource for scholars studying prejudices against vulnerable groups~\cite{saha2019prevalence}, our work also contributes to the development of fair hate speech detection systems. Our work is relevant in this area since human biases hinder the manual annotation of training and testing datasets. Moreover, LLMs can be used for automatic data annotation, data augmentation, or directly for hate speech detection. Thus, their biases may also cause multiple negative downstream consequences. Our work enhances the understanding of these biases and informs the construction, curation, and rebalancing of hate speech datasets, as well as the development of new methods for training fair hate speech detectors~\cite{garg2023handling}.\hl{\\Our study also serves as a foundational tool to inform broader discussions on normative, ethical, and practical questions regarding LLM behavior. By providing empirical insights into the biases present in human annotators and persona-based LLMs during hate speech detection, we aim to contribute to the larger discourse. These findings can help guide future research and decision-making in addressing critical issues, such as whether LLMs should aim for unbiased outputs or mirror human decision-making in sensitive contexts.} \hl{ Additionally, our findings have practical implications for platform moderation policies and corporate AI ethics. By identifying socio-demographic biases in hate speech annotation, platforms can refine guidelines and training to ensure fair content moderation, which limits the risk of user abandonment~\cite{cima2024investigating, tessa2024beyond}. Similarly, corporate AI ethics frameworks can leverage these insights to develop and deploy more equitable AI systems, reducing the risk of perpetuating stereotypes and enhancing fairness in automated decision-making.} 

\textbf{Generalizability.} While our present study focuses on bias in hate speech annotations, our approach is general and broadly applicable. The methods we employed to uncover and characterize human biases can be used to detect biases in a wide range of annotated datasets. Additionally, our methodology for personalizing LLMs and comparing their biases to human ones can be easily carried over to other tasks beyond hate speech detection. 

\textbf{Limitations and Future work.} Some limitations must be acknowledged. \hl{While our methodology focuses on capturing relative differences in labeling behaviors between groups with and without specific attributes, we recognize that demographic imbalances in annotator and target populations may influence the observed patterns of bias. For example, certain annotator groups may be underrepresented, leading to a less comprehensive understanding of how their socio-demographic characteristics affect labeling decisions. Similarly, the overrepresentation of certain target groups as recipients of hate speech could amplify or obscure specific biases. Although our analysis reflects the data as collected, addressing these imbalances through stratified sampling or additional weighting strategies could further disentangle these effects in future work. Another possible limitation is that the social media data, although extensive and rich, was collected from platforms with a marked US user-base and the annotations were obtained from US crowdsourcing contributors. Hence the data may primarily reflect US language, cultural norms, and perspectives, potentially overlooking linguistic practices and cultural contexts from other regions or communities. This limitation may affect the applicability of the findings to other linguistic or geographic contexts. Future research should consider conducting similar experiments in different languages and cultural settings to enable meaningful comparisons with the results obtained in this study.} Methodologically, our analysis of annotation bias, while informative, does not capture all dimensions of bias, such as contextual factors or annotators' subjective experiences. Moreover, although we experimented with various prompts and LLM models, the specific prompting strategy and LLM versions used may have influenced the results. These limitations highlight the need for more representative datasets, improved methods for LLM personalization such as automatic prompt optimization~\cite{schulhoff2024prompt}, and deeper investigations into specific biases in future research. Other than these, future work should also focus on intersectional studies, as combinations of attributes, rather than individual ones, can cause new biases or reinforce existing ones~\cite{kim2020intersectional}. To this end, the analyzed dataset appears particularly well-suited for comprehensive intersectional analyses given the availability of rich socio-demographic information for both annotators and targets. \section{Acknowledgments}
This work is partially supported by the European Union -- Next Generation EU within the PRIN 2022 project \texttt{PIANO} (Personalized Interventions Against Online Toxicity), and within the ERC project DEDUCE (\textit{Data-driven and User-centered Content Moderation}) under grant \#101113826; by the PNRR-M4C2 (PE00000013) “FAIR-Future Artificial Intelligence Research" - Spoke 1 "Human-centered AI", funded under Next Generation EU; and by the Italian Ministry of Education and Research (MUR) in the framework of the FoReLab projects (Departments of Excellence).
 
\bibliography{references} 

\begin{thebibliography}{44}
\providecommand{\natexlab}[1]{#1}

\bibitem[{Abrate, Bacciu, and Cima(2024)}]{Abrate2024}
Abrate, M.; Bacciu, C.; and Cima, L. 2024.
\newblock Visualizing Hate Speech Biases -- Companion visualization.
\newblock doi.org/10.6084/m9.figshare.27220917.v1.

\bibitem[{Agiza, Mostagir, and Reda(2024)}]{agiza2024politune}
Agiza, A.; Mostagir, M.; and Reda, S. 2024.
\newblock {PoliTune}: Analyzing the impact of data selection and fine-tuning on
  economic and political biases in large language models.
\newblock In \emph{AAAI/ACM AIES}.

\bibitem[{Aher, Arriaga, and Kalai(2023)}]{aher2023using}
Aher, G.~V.; Arriaga, R.~I.; and Kalai, A.~T. 2023.
\newblock Using large language models to simulate multiple humans and replicate
  human subject studies.
\newblock In \emph{ICML}.

\bibitem[{Al~Kuwatly, Wich, and Groh(2020)}]{al2020identifying}
Al~Kuwatly, H.; Wich, M.; and Groh, G. 2020.
\newblock Identifying and measuring annotator bias based on annotators’
  demographic characteristics.
\newblock In \emph{WOAH}.

\bibitem[{Argyle et~al.(2023)Argyle, Busby, Fulda, Gubler, Rytting, and
  Wingate}]{argyle2023out}
Argyle, L.~P.; Busby, E.~C.; Fulda, N.; Gubler, J.~R.; Rytting, C.; and
  Wingate, D. 2023.
\newblock Out of one, many: Using language models to simulate human samples.
\newblock \emph{Political Analysis}, 31(3).

\bibitem[{Baack(2024)}]{baack2024critical}
Baack, S. 2024.
\newblock A critical analysis of the largest source for generative AI training
  data: Common Crawl.
\newblock In \emph{ACM FAccT}.

\bibitem[{Beck et~al.(2024)Beck, Schuff, Lauscher, and
  Gurevych}]{beck2024sensitivity}
Beck, T.; Schuff, H.; Lauscher, A.; and Gurevych, I. 2024.
\newblock Sensitivity, performance, robustness: Deconstructing the effect of
  sociodemographic prompting.
\newblock In \emph{EACL}.

\bibitem[{Cima et~al.(2024{\natexlab{a}})Cima, Miaschi, Trujillo, Avvenuti,
  Dell'Orletta, and Cresci}]{cima2024contextualized}
Cima, L.; Miaschi, A.; Trujillo, A.; Avvenuti, M.; Dell'Orletta, F.; and
  Cresci, S. 2024{\natexlab{a}}.
\newblock {Contextualized counterspeech: Strategies for adaptation,
  personalization, and evaluation}.
\newblock In \emph{ACM Web Conference}.

\bibitem[{Cima et~al.(2024{\natexlab{b}})Cima, Tessa, Cresci, Trujillo, and
  Avvenuti}]{cima2024investigating}
Cima, L.; Tessa, B.; Cresci, S.; Trujillo, A.; and Avvenuti, M.
  2024{\natexlab{b}}.
\newblock Investigating the heterogeneous effects of a massive content
  moderation intervention via Difference-in-Differences.
\newblock \emph{arXiv:2411.04037}.

\bibitem[{Das et~al.(2024)Das, Zhang, Jamshidi, Jain, Chadha, Raychawdhary,
  Sandage, Pope, Dozier, and Seals}]{das2024investigating}
Das, A.; Zhang, Z.; Jamshidi, F.; Jain, V.; Chadha, A.; Raychawdhary, N.;
  Sandage, M.; Pope, L.; Dozier, G.; and Seals, C. 2024.
\newblock Investigating annotator bias in large language models for hate speech
  detection.
\newblock \emph{arXiv:2406.11109}.

\bibitem[{Davani et~al.(2023)Davani, Atari, Kennedy, and
  Dehghani}]{davani2023hate}
Davani, A.~M.; Atari, M.; Kennedy, B.; and Dehghani, M. 2023.
\newblock Hate speech classifiers learn normative social stereotypes.
\newblock \emph{TACL}, 11.

\bibitem[{Garg et~al.(2023)Garg, Masud, Suresh, and
  Chakraborty}]{garg2023handling}
Garg, T.; Masud, S.; Suresh, T.; and Chakraborty, T. 2023.
\newblock Handling bias in toxic speech detection: A survey.
\newblock \emph{ACM CSUR}.

\bibitem[{Geva, Goldberg, and Berant(2019)}]{geva2019we}
Geva, M.; Goldberg, Y.; and Berant, J. 2019.
\newblock Are we modeling the task or the annotator? An investigation of
  annotator bias in natural language understanding datasets.
\newblock In \emph{EMNLP-IJCNLP}.

\bibitem[{Gupta et~al.(2024)Gupta, Shrivastava, Deshpande, Kalyan, Clark,
  Sabharwal, and Khot}]{gupta2023bias}
Gupta, S.; Shrivastava, V.; Deshpande, A.; Kalyan, A.; Clark, P.; Sabharwal,
  A.; and Khot, T. 2024.
\newblock Bias runs deep: Implicit reasoning biases in persona-assigned LLMs.
\newblock In \emph{ICLR}.

\bibitem[{Hettiachchi et~al.(2023)Hettiachchi, Holcombe-James, Livingstone,
  de~Silva, Lease, Salim, and Sanderson}]{hettiachchi2023crowd}
Hettiachchi, D.; Holcombe-James, I.; Livingstone, S.; de~Silva, A.; Lease, M.;
  Salim, F.~D.; and Sanderson, M. 2023.
\newblock How crowd worker factors influence subjective annotations: A study of
  tagging misogynistic hate speech in tweets.
\newblock In \emph{AAAI HCOMP}.

\bibitem[{Hu and Collier(2024)}]{hu2024quantifying}
Hu, T.; and Collier, N. 2024.
\newblock Quantifying the persona effect in LLM simulations.
\newblock In \emph{ACL}.

\bibitem[{Hu et~al.(2024)Hu, Kyrychenko, Rathje, Collier, van~der Linden, and
  Roozenbeek}]{hu2024generative}
Hu, T.; Kyrychenko, Y.; Rathje, S.; Collier, N.; van~der Linden, S.; and
  Roozenbeek, J. 2024.
\newblock Generative language models exhibit social identity biases.
\newblock \emph{Nature Computational Science}.

\bibitem[{Kim et~al.(2020)Kim, Ortiz, Nam, Santiago, and
  Datta}]{kim2020intersectional}
Kim, J.~Y.; Ortiz, C.; Nam, S.; Santiago, S.; and Datta, V. 2020.
\newblock Intersectional bias in hate speech and abusive language datasets.
\newblock In \emph{AAAI ICWSM}.

\bibitem[{La~Cava and Tagarelli(2025)}]{la2025open}
La~Cava, L.; and Tagarelli, A. 2025.
\newblock Open models, closed minds? on agents capabilities in mimicking human
  personalities through open large language models.
\newblock In \emph{AAAI Conference on Artificial Intelligence}, volume~39,
  1355--1363.

\bibitem[{Lee, An, and Thorne(2023)}]{lee2023can}
Lee, N.; An, N.~M.; and Thorne, J. 2023.
\newblock Can large language models capture dissenting human voices?
\newblock In \emph{EMNLP}.

\bibitem[{Nogara et~al.(2025)Nogara, Pierri, Cresci, Luceri, T\"{o}rnberg, and
  Giordano}]{nogara2024toxic}
Nogara, G.; Pierri, F.; Cresci, S.; Luceri, L.; T\"{o}rnberg, P.; and Giordano,
  S. 2025.
\newblock {Toxic bias: Perspective API misreads German as more toxic}.
\newblock In \emph{AAAI ICWSM}.

\bibitem[{Ouyang et~al.(2022)Ouyang, Wu, Jiang, Almeida, Wainwright, Mishkin,
  Zhang, Agarwal, Slama, Ray et~al.}]{ouyang2022training}
Ouyang, L.; Wu, J.; Jiang, X.; Almeida, D.; Wainwright, C.; Mishkin, P.; Zhang,
  C.; Agarwal, S.; Slama, K.; Ray, A.; et~al. 2022.
\newblock Training language models to follow instructions with human feedback.
\newblock \emph{NeurIPS}.

\bibitem[{Park et~al.(2024)Park, Zou, Shaw, Hill, Cai, Morris, Willer, Liang,
  and Bernstein}]{park2024generative}
Park, J.~S.; Zou, C.~Q.; Shaw, A.; Hill, B.~M.; Cai, C.; Morris, M.~R.; Willer,
  R.; Liang, P.; and Bernstein, M.~S. 2024.
\newblock Generative agent simulations of 1,000 people.
\newblock \emph{arXiv:2411.10109}.

\bibitem[{Parmar et~al.(2023)Parmar, Mishra, Geva, and Baral}]{parmar2023don}
Parmar, M.; Mishra, S.; Geva, M.; and Baral, C. 2023.
\newblock Don’t blame the annotator: Bias already starts in the annotation
  instructions.
\newblock In \emph{EACL}.

\bibitem[{Perez et~al.(2023)Perez, Ringer, Lukosiute, Nguyen, Chen, Heiner,
  Pettit, Olsson, Kundu, Kadavath et~al.}]{perez2023discovering}
Perez, E.; Ringer, S.; Lukosiute, K.; Nguyen, K.; Chen, E.; Heiner, S.; Pettit,
  C.; Olsson, C.; Kundu, S.; Kadavath, S.; et~al. 2023.
\newblock Discovering language model behaviors with model-written evaluations.
\newblock In \emph{ACL}.

\bibitem[{Prabhakaran, Davani, and D{\'\i}az(2021)}]{prabhakaran2021releasing}
Prabhakaran, V.; Davani, A.~M.; and D{\'\i}az, M. 2021.
\newblock On releasing annotator-level labels and information in datasets.
\newblock In \emph{LAW-DMR}.

\bibitem[{Rao, Leung, and Miao(2023)}]{rao2023can}
Rao, H.; Leung, C.; and Miao, C. 2023.
\newblock Can ChatGPT Assess Human Personalities? A General Evaluation
  Framework.
\newblock In \emph{EMNLP}.

\bibitem[{Rossetti et~al.(2024)Rossetti, Stella, Cazabet, Abramski, Cau,
  Citraro, Failla, Improta, Morini, and Pansanella}]{rossetti2024social}
Rossetti, G.; Stella, M.; Cazabet, R.; Abramski, K.; Cau, E.; Citraro, S.;
  Failla, A.; Improta, R.; Morini, V.; and Pansanella, V. 2024.
\newblock {Y Social}: An {LLM}-powered social media digital twin.
\newblock \emph{arXiv:2408.00818}.

\bibitem[{Sachdeva et~al.(2022{\natexlab{a}})Sachdeva, Barreto, Bacon, Sahn,
  Von~Vacano, and Kennedy}]{sachdeva2022measuring}
Sachdeva, P.; Barreto, R.; Bacon, G.; Sahn, A.; Von~Vacano, C.; and Kennedy, C.
  2022{\natexlab{a}}.
\newblock The measuring hate speech corpus: Leveraging Rasch measurement theory
  for data perspectivism.
\newblock In \emph{NLPerspectives}.

\bibitem[{Sachdeva et~al.(2022{\natexlab{b}})Sachdeva, Barreto, von Vacano, and
  Kennedy}]{sachdeva2022assessing}
Sachdeva, P.~S.; Barreto, R.; von Vacano, C.; and Kennedy, C.~J.
  2022{\natexlab{b}}.
\newblock Assessing annotator identity sensitivity via item response theory: A
  case study in a hate speech corpus.
\newblock In \emph{ACM FAccT}.

\bibitem[{Saha, Chandrasekharan, and De~Choudhury(2019)}]{saha2019prevalence}
Saha, K.; Chandrasekharan, E.; and De~Choudhury, M. 2019.
\newblock Prevalence and psychological effects of hateful speech in online
  college communities.
\newblock In \emph{ACM WebSci}.

\bibitem[{Santurkar et~al.(2023)Santurkar, Durmus, Ladhak, Lee, Liang, and
  Hashimoto}]{santurkar2023whose}
Santurkar, S.; Durmus, E.; Ladhak, F.; Lee, C.; Liang, P.; and Hashimoto, T.
  2023.
\newblock Whose opinions do language models reflect?
\newblock In \emph{ICML}.

\bibitem[{Sap et~al.(2022)Sap, Swayamdipta, Vianna, Zhou, Choi, and
  Smith}]{sap2022annotators}
Sap, M.; Swayamdipta, S.; Vianna, L.; Zhou, X.; Choi, Y.; and Smith, N.~A.
  2022.
\newblock Annotators with attitudes: How annotator beliefs and identities bias
  toxic language detection.
\newblock In \emph{NAACL-HLT}.

\bibitem[{Sch{\"a}fer et~al.(2024)Sch{\"a}fer, Combs, Bagdon, Li, Probol,
  Greschner, Papay, Resendiz, Velutharambath, W{\"u}hrl
  et~al.}]{schafer2024demographics}
Sch{\"a}fer, J.; Combs, A.; Bagdon, C.; Li, J.; Probol, N.; Greschner, L.;
  Papay, S.; Resendiz, Y.~M.; Velutharambath, A.; W{\"u}hrl, A.; et~al. 2024.
\newblock Which demographics do {LLMs} default to during annotation?
\newblock \emph{arXiv:2410.08820}.

\bibitem[{Schulhoff et~al.(2024)Schulhoff, Ilie, Balepur, Kahadze, Liu, Si, Li,
  Gupta, Han, Schulhoff et~al.}]{schulhoff2024prompt}
Schulhoff, S.; Ilie, M.; Balepur, N.; Kahadze, K.; Liu, A.; Si, C.; Li, Y.;
  Gupta, A.; Han, H.; Schulhoff, S.; et~al. 2024.
\newblock The Prompt Report: A systematic survey of prompting techniques.
\newblock \emph{arXiv:2406.06608}.

\bibitem[{Simmons(2023)}]{simmons2023moral}
Simmons, G. 2023.
\newblock Moral mimicry: Large language models produce moral rationalizations
  tailored to political identity.
\newblock In \emph{ACL}.

\bibitem[{Srivastava et~al.(2023)Srivastava, Rastogi, Rao, Shoeb, Abid, Fisch
  et~al.}]{srivastava2023beyond}
Srivastava, A.; Rastogi, A.; Rao, A.; Shoeb, A. A.~M.; Abid, A.; Fisch, A.;
  et~al. 2023.
\newblock Beyond the Imitation Game: Quantifying and extrapolating the
  capabilities of language models.
\newblock \emph{TMLR}.

\bibitem[{Talat and Hovy(2016)}]{talat2016hateful}
Talat, Z.; and Hovy, D. 2016.
\newblock Hateful symbols or hateful people? predictive features for hate
  speech detection on twitter.
\newblock In \emph{Proceedings of the NAACL}, 88--93.

\bibitem[{Tessa et~al.(2024)Tessa, Cima, Trujillo, Avvenuti, and
  Cresci}]{tessa2024beyond}
Tessa, B.; Cima, L.; Trujillo, A.; Avvenuti, M.; and Cresci, S. 2024.
\newblock Beyond Trial-and-Error: Predicting User Abandonment After a
  Moderation Intervention.
\newblock \emph{arXiv:2404.14846}.

\bibitem[{T{\"o}rnberg et~al.(2023)T{\"o}rnberg, Valeeva, Uitermark, and
  Bail}]{tornberg2023simulating}
T{\"o}rnberg, P.; Valeeva, D.; Uitermark, J.; and Bail, C. 2023.
\newblock Simulating social media using large language models to evaluate
  alternative news feed algorithms.
\newblock \emph{arXiv:2310.05984}.

\bibitem[{Tseng et~al.(2024)Tseng, Huang, Hsiao, Chen, Huang, Meng, and
  Chen}]{tseng2024two}
Tseng, Y.-M.; Huang, Y.-C.; Hsiao, T.-Y.; Chen, W.-L.; Huang, C.-W.; Meng, Y.;
  and Chen, Y.-N. 2024.
\newblock Two Tales of Persona in LLMs: A Survey of Role-Playing and
  Personalization.
\newblock In \emph{Findings of the Association for Computational Linguistics:
  EMNLP 2024}, 16612--16631.

\bibitem[{Wan et~al.(2023)Wan, Wang, He, Gu, Bai, and Lyu}]{wan2023biasasker}
Wan, Y.; Wang, W.; He, P.; Gu, J.; Bai, H.; and Lyu, M.~R. 2023.
\newblock BiasAsker: Measuring the bias in conversational AI system.
\newblock In \emph{ACM ESEC/FSE}.

\bibitem[{Wich, Al~Kuwatly, and Groh(2020)}]{wich2020investigating}
Wich, M.; Al~Kuwatly, H.; and Groh, G. 2020.
\newblock Investigating annotator bias with a graph-based approach.
\newblock In \emph{WOAH}.

\bibitem[{Wich et~al.(2021)Wich, Widmer, Hagerer, and
  Groh}]{wich2021investigating}
Wich, M.; Widmer, C.; Hagerer, G.; and Groh, G. 2021.
\newblock Investigating annotator bias in abusive language datasets.
\newblock In \emph{RANLP}.

\end{thebibliography}
\newpage
\appendix

\renewcommand{\thetable}{T\arabic{table}}
\renewcommand{\thefigure}{F\arabic{figure}}
\renewcommand{\thealgorithm}{A\arabic{algorithm}}
\setcounter{table}{0}
\setcounter{figure}{0}
\setcounter{algorithm}{0}

\section{Appendix}

\clearpage{}\begin{table*}[t]
\footnotesize
\setlength{\tabcolsep}{2pt}
    \centering
\hspace{-0.04\textwidth}\begin{minipage}[b]{0.45\textwidth}\begin{tabular}{clrcrccc}
	\toprule
        \textbf{attribute} & \textbf{value} & \textbf{annotators} && \textbf{targets}\\
    \midrule
        \textit{age} & young adults & 4,964 (62.78\%) && 935 (26.55\%) \\
        & teenagers & 201 (2.54\%) && 786 (22.32\%) \\
        \rowcolor{gray!30!} & children & 0 (0\%) && 642 (18.23\%) \\
        & middle aged & 2,306 (29.16\%) && 560 (15.90\%) \\
        & seniors & 436 (5.52\%) && 500 (14.19\%) \\
        \rowcolor{gray!30!} & other & - && 99 (2.81\%) \\
    \midrule
        \textit{gender} & women & 4,426 (55.67\%) && 27,889 (54.34\%) \\
        & men & 3,392 (42.67\%) && 10,029 (19.54\%) \\
        & transgender unspecified & 67 (0.84\%) && 4,703 (9.16\%) \\
        & transgender men & - && 3,326 (6.48\%) \\
        & transgender women & - && 2,611 (5.09\%) \\
        & non binary & 59 (0.74\%) && 2,116 (4.12\%) \\
        \rowcolor{gray!30!} & other & 6 (0.08\%) && 651 (1.27\%) \\
    \midrule
        \textit{race} & black & 791 (9.21\%) && 22,899 (33.86\%) \\
        & white & 6,373 (74.2\%) && 9,797 (14.49\%) \\
        & middle eastern & 51 (0.59\%) && 9,450 (13.97\%) \\
        & latinx & 560 (6.52\%) && 8,497 (12.56\%) \\
        & asian & 552 (6.43\%) && 7,025 (10.39\%) \\
        & native american & 154 (1.79\%) && 2,819 (4.17\%) \\
        & pacific islander & 28 (0.33\%) && 2,358 (3.49\%) \\
        \rowcolor{gray!30!} & other & 80 (0.93\%) && 4,780 (7.07\%) \\
    \midrule
       \textit{religion} & muslim & 56 (0.70\%) && 12,509 (38.30\%) \\ 
        & christian & 3,350 (41.89\%) && 6,982 (21.38\%) \\
        & jewish & 128 (1.60\%) && 6,924 (21.20\%) \\
        & hindu & 37 (0.46\%) && 1,285 (3.93\%) \\
        & atheist & 1,608 (20.11\%) && 953 (2.92\%) \\
        & mormon & 60 (0.75\%) && 953 (2.92\%) \\
        & buddhist & 126 (1.58\%) && 729 (2.23\%) \\
        \rowcolor{gray!30!} & other & 497 (6.21\%) && 2,328 (7.12\%) \\
        \rowcolor{gray!30!} & nothing & 2,191 (27.40\%) && - \\
    \midrule
        \textit{sexuality} & gay & 308 (3.91\%) && 15,465 (42.34\%) \\   
        & lesbian & - && 6,883 (18.85\%) \\
        & bisexual & 727 (9.22\%) && 6,631 (18.16\%) \\
        & straight & 6,727 (85.30\%) && 4,438 (12.15\%) \\
        \rowcolor{gray!30!} & other & 124 (1.57\%) && 3,104 (8.50\%) \\
    \bottomrule  
    \end{tabular}
    \end{minipage}\hspace{0.07\textwidth}\begin{minipage}[b]{0.45\textwidth}\begin{tabular}{llrcrc}
	\toprule
        \textbf{attribute} & \textbf{value} & \textbf{annotators} && \textbf{targets} \\
        \midrule
    \textit{disability} & cognitive & - && 1,735 (36.77\%) \\    
        & physical & - && 1,126 (23.86\%) \\
        & unspecific & - && 855 (18.12\%) \\
        & neurological & - && 542 (11.48\%) \\
        & hearing impaired & - && 142 (3.01\%) \\
        & visually impaired & - && 125 (2.65\%) \\
        \rowcolor{gray!30!} & other & - && 194 (4.11\%) \\ 
    \midrule
        \textit{education} & college ba & 2,913 (36.82\%) && - \\     
        & some college & 2,064 (26.09\%) && - \\
        & college aa & 1,041 (13.16\%) && - \\
        & high school & 850 (10.74\%) && - \\
        & master & 729 (9.22\%) && - \\
        & professional degree & 176 (2.22\%) && - \\
        & PhD & 89 (1.13\%) && - \\
        & some high school & 49 (0.62\%) && - \\
     \midrule
        \textit{ideology} & liberal & 1,968 (24.88\%) && - \\      
        & neutral & 1,367 (17.28\%) && - \\
        & slightly liberal & 1,242 (15.70\%) && - \\
        \rowcolor{gray!30!} & extremely liberal & 1,065 (13.46\%) && - \\
        & conservative & 895 (11.31\%) && - \\
        & slightly conservative & 860 (10.87\%) && - \\
        & extremely conservative & 264 (3.34\%) && - \\
        & no opinion & 249 (3.16\%) && - \\
    \midrule
        \textit{income} & 10K - 50K & 3,321 (42.01\%) && - \\      
        & 50K - 100K & 3,079 (38.94\%) && - \\
        & 100K - 200K & 1,009 (12.76\%) && - \\
        & $<$ 10K & 375 (4.74\%) && - \\
        & $>$ 200K & 122 (1.55\%) && - \\
    \midrule
        \textit{origin} & specific country & - && 14,124 (43.45\%) \\
        & immigrant & - && 9,525 (29.30\%) \\
        & undocumented & - && 6,081 (18.71\%) \\
        & migrant worker & - && 2,523 (7.76\%) \\
        \rowcolor{gray!30!} & other & - && 250 (0.78\%) \\
    \bottomrule    
    \end{tabular}
    \end{minipage}
\caption{The dataset includes socio-demographic attributes for both annotators and hate targets, with each attribute taking on multiple values. In this comprehensive overview, we present the distributions of these values. The left column reports attributes that are available for both annotators and hate targets, while the right columns displays attributes specific to only one of these categories. Grey-colored rows represent attributes that are excluded from the results, either because they did not yield statistically significant findings (e.g., \textbf{ideology:} \textit{extremely liberal}) or because they were not used, such as all \textit{other} values.}
\label{tab:dataset-large}
\end{table*}
\clearpage{}

\begin{algorithm*}[]
\caption{Pseudo-code to compute a confusion matrix $D$ for annotators having the socio-demographic attribute $\mathbf{t} = v$, annotating posts targeted at individuals with attribute $\mathbf{t'} = v'$.} \begin{algorithmic}[1]
\small
\Function{compute\_matrix}{$\mathbf{t}=v$, $\mathbf{t'}=v'$}
    \State $D \gets$ \Call{init\_matrix}{} \Comment{Init matrix with zeros}
    \ForAll{$comment \in G(\mathbf{t'}=v')$}
        \ForAll{$annotator \in A(\mathbf{t}=v)$}
            \State $text\_analyzed \gets$ \Call{extract\_text}{comment} \Comment{Get the text associated with current comment record}
            \State $other\_comments \gets$ \Call{get\_records\_from\_text}{text\_analyzed} \Comment{Get all records sharing this text}
            \ForAll{$other\_comment \in other\_comments$}
                \If{$comment = other\_comment$}
                    \State \textbf{continue}
                \EndIf
                \State $An \gets$ \Call{annotator}{comment} \Comment{Get annotator attribute/value}
                \State $An\_other \gets$ \Call{annotator}{other\_comment} \Comment{Get annotator attribute/value}
                \If{$An \neq An\_other$} 
                    \State $comment\_label \gets$ \Call{get\_hatespeech\_label}{comment} \Comment{T, N, or M}
                    \State $other\_comment\_label \gets$ \Call{get\_hatespeech\_label}{other\_comment} \Comment{T, N, or M}
                    \State \Call{inc\_cell\_matrix}{D, comment\_label, other\_comment\_label, 1} \Comment{Inc by 1 corresponding matrix cell}
                \EndIf
            \EndFor
        \EndFor
    \EndFor
    \State \Return $D$
\EndFunction
\end{algorithmic}
\label{Alg:ConfusionMatrix}
\end{algorithm*}

\begin{table*}[t]
\centering
\begin{tabular}{p{0.95\textwidth}}
\small
\begin{tcolorbox}[fontupper=\small, coltext=black, width=0.95\textwidth, halign=justify]
Objective: You are an experienced human annotator who must analyze a social network text comment and determine whether the comment targets a protected group and determine if it contains hate speech or not.\\
To state if the text comment contains hate speech, you must answer the following question:  \newline \newline 
Does this comment contain hate speech, defined as ``bias-motivated, hostile and malicious language targeted at a person/group because of their actual or perceived innate characteristics, especially when the group is unnecessarily labeled (EXAMPLE: ``jew bankers'').''
\newline \newline
Also, you must take in mind that:
\newline \newline
The comment contains hate speech if slurs or derogatory language are used toward any person or group regardless of whether or not the target is a member of the slur-target group. (e.g. a comment calling Trump a faggot).
\newline \newline
If you can't decide if the comment contains hate speech, you must label it as ``maybe''
\newline \newline
To determine the comment targets, you must select at least one target from the following list:
\{target\_list\}
\newline \newline
In performing the task, you must impersonate an annotator with the following listed characteristics:
\begin{itemize}
\setlength\itemsep{0em}
    \item annotator\_gender: \{annotator\_gender\}
    \item \{annotator\_trans\}
    \item \{annotator\_sexuality\}
    \item \{annotator\_race\}
    \item \{annotator\_religion\}
    \item annotator\_education: \{annotator\_educ\}
    \item annotator\_income: \{annotator\_income\} 
    \item annotator\_ideology: \{annotator\_ideology\}
    \item annotator\_age: \{annotator\_age\} years.
\end{itemize}
\vspace{0.2cm}
You must analyze the input text and predict an output in JSON format that indicates: 
\begin{itemize}
\setlength\itemsep{0em}
    \item hate\_speech: (yes|no|maybe)
    \item A list named 'targets' including at least 1 or more targets listed above with hate speech expressed in the text.
    \item 'targets' list must be not empty
\end{itemize}
You must only return raw JSON output as described above, in particular:
\begin{itemize}
\setlength\itemsep{0em}
    \item don't generate any explanations or additional notes of the reasons of such responses
    \item don't escape any character
\end{itemize}
The input text to analyze is:
\newline \newline
\{comment\_text\}
\end{tcolorbox}
\end{tabular}
\caption{Prompt template used for dynamically personalizing LLMs with human annotators' socio-demographic characteristics and for annotating hate speech messages.}
\label{tab:llm-prompt}
\end{table*}
 \begin{table*}[t]
\centering
\begin{tabular}{p{0.95\textwidth}}
\begin{tcolorbox}[fontupper=\small, coltext=black, width=0.95\textwidth, halign=justify]
Objective: You are an experienced human annotator who must analyze a social network text comment and determine whether the comment targets a protected group and determine if it contains hate speech or not.\\
To state if the text comment contains hatespeech, you must answer the following question:  \newline \newline 
Does this comment contain hate speech, defined as "bias-motivated, hostile and malicious language targeted at a person/group because of their actual or perceived innate characteristics, especially when the group is unnecessarily labeled (EXAMPLE: "jew bankers")."
\newline \newline
Also, you must take in mind that:
\newline \newline
The comment contains hate speech if slurs or derogatory language are used toward any person or group regardless of whether or not the target is a member of the slur-target group. (e.g. a comment calling Trump a faggot).
\newline \newline
If you can't decide if the comment contains hate speech, you must label it as "maybe"
\newline \newline
To determine the comment targets, you must select at least one target from the following list:
\{target\_list\}
\newline \newline
In performing the task, you must impersonate an annotator with the following feature:
\begin{itemize}
\setlength\itemsep{0em}
    \item annotator\_feature: \{annotator\_random\_feature\}
\end{itemize}
\vspace{0.2cm}
You must analyze the input text and predict an output in JSON format that indicates: 
\begin{itemize}
\setlength\itemsep{0em}
    \item hate\_speech: (yes|no|maybe)
    \item A list named 'targets' including at least 1 or more targets listed above with hate speech expressed in the text.
    \item 'targets' list must be not empty
\end{itemize}
You must only return raw JSON output as described above, in particular:
\begin{itemize}
\setlength\itemsep{0em}
    \item don't generate any explanations or additional notes of the reasons of such responses
    \item don't escape any character
\end{itemize}
The input text to analyze is:
\newline \newline
\{comment\_text\}
\end{tcolorbox}
\end{tabular}
\caption{Alternative prompt obtained by reducing the number of socio-demographic attributes passed to the model.}
\label{tab:llm-prompt-2}
\end{table*}
 \begin{table*}[t]
\centering
\begin{tabular}{p{0.95\textwidth}}
\begin{tcolorbox}[fontupper=\small, coltext=black, width=0.95\textwidth, halign=justify]
Objective: You are an experienced annotator tasked with analyzing a social network comment to identify if it targets a protected group and contains hate speech.
\newline \newline
Hate speech is defined as "bias-motivated, hostile, and malicious language aimed at a person/group based on their actual or perceived innate characteristics, especially when the group is unnecessarily labeled (e.g., 'jew bankers')."
\newline \newline
A comment qualifies as hate speech if slurs or derogatory language are used, even if the target is not a member of the slur-targeted group (e.g., calling Trump a faggot). If unsure, label it as "maybe."
\newline \newline
To determine the targets, select at least one from the following list:
\{target\_list\}
\newline \newline
In performing the task, you must impersonate an annotator with the following listed characteristics:
\begin{itemize}
\setlength\itemsep{0em}
    \item annotator\_gender: \{annotator\_gender\}
    \item \{annotator\_trans\}
    \item \{annotator\_sexuality\}
    \item \{annotator\_race\}
    \item \{annotator\_religion\}
    \item annotator\_education: \{annotator\_educ\}
    \item annotator\_income: \{annotator\_income\} 
    \item annotator\_ideology: \{annotator\_ideology\}
    \item annotator\_age: \{annotator\_age\} years.
\end{itemize}
\vspace{0.2cm}
You must predict and return the following output in JSON format:
\begin{itemize}
\setlength\itemsep{0em}
    \item hate\_speech: (yes|no|maybe)
    \item A list named 'targets' containing at least one target from the above list.
\end{itemize}
Return only the raw JSON output with no explanations or notes.
\newline \newline
The input text to analyze is:
\newline \newline
\{comment\_text\}
\end{tcolorbox}
\end{tabular}
\caption{Alternative prompt obtained by simplifying the instructions for the hate speech annotation task.}
\label{tab:llm-prompt-3}
\end{table*}

\end{document}